\documentclass[a4paper,fleqn]{cas-dc}

\usepackage[numbers,sort]{natbib}
\usepackage{float}

\usepackage{epstopdf}
\usepackage{color}
\usepackage[noend]{algpseudocode}
\usepackage{algorithmicx,algorithm}

\usepackage{graphicx}
\usepackage{subcaption}

\usepackage{booktabs}       
\usepackage{amsfonts}       
\usepackage{nicefrac}       
\usepackage[nopatch=footnote]{microtype}  
\usepackage{xcolor}         
\usepackage{graphicx}
\usepackage{wrapfig}
\usepackage{multirow}
\usepackage{epsfig}
\usepackage{amsmath}
\usepackage{amssymb}
\usepackage{graphicx}

\let\savedopenbox\openbox
\usepackage{newtxtext,newtxmath}  
\let\openbox\savedopenbox
\usepackage{nomencl} 
\usepackage{tcolorbox} 
\usepackage{multicol} 
\usepackage{threeparttable} 
\usepackage{tabularx} 
\makenomenclature 

\nomenclature{WECN}{Wavelet-Enhanced Convolutional Network}

\nomenclature{TCS}{Tidal Current Speed}
\nomenclature{TCSF}{Tidal Current Speed Forecasting}
\nomenclature{FFT}{Fast Fourier Transform}
\nomenclature{DWT}{Discrete Wavelet Transform}
\nomenclature{TPE}{Tree-structured Parzen Estimator}
\nomenclature{MAE}{Mean Absolute Error}
\nomenclature{MSE}{Mean Squared Error}
\nomenclature{MAPE}{Mean Absolute Percentage Error}
\nomenclature{MSPE}{Mean Squared Percentage Error}
\nomenclature{BSL}{Below Sea Level}
\nomenclature{RS}{Random Sampler}
\nomenclature{$\textbf{\textit{X}}_{\text{1D}}$}{Multivariate input time series}
\nomenclature{$\textbf{\textit{x}}^n(t)$}{The time point on the \(n\)-th sequence in $\textbf{\textit{X}}_{\text{1D}}$}
\nomenclature{$\textbf{\textit{cD}}_j$}{Average detail coefficients at the \(j\)-th level for all sequences}
\nomenclature{$\xi_j$}{Average amplitude at the \(j\)-th level}
\nomenclature{$k$}{Top \(k\) amplitudes are selected
}
\nomenclature{$T$}{Number of sampling points per sequence}
\nomenclature{$N$}{Number of sequences}
\nomenclature{$p_i$}{The \(i\)-th period}
\nomenclature{$f_i$}{The \(i\)-th frequency}
\nomenclature{$A_{f_i}$}{Amplitude corresponding to \(f_i\)}
\nomenclature{$\textbf{\textit{X}}_{\text{1D}}^l$}{Features of the \(l\)-th layer of \(X_{\text{1D}}\)}
\nomenclature{$\textbf{\textit{X}}_{\text{2D}}^{l, i}$}{2D tensor obtained by reshaping}
\nomenclature{$\hat{\mathbf{X}}^{l, i}_{\text{2D}}$}{2D tensor processed by Inception}
\nomenclature{$\hat{\mathbf{X}}^{l,i}_{\text{1D}}$}{1D sequence after reshaping back}
\nomenclature{$\hat{A}^{l-1}_{f_1}$}{Weights for sequence aggregation}
\nomenclature{$\beta$}{Output length}
\nomenclature{$\hat{\textbf{\textit{x}}}^{n}(t)$}{The forecasted value}
\nomenclature{$y$}{Loss function}
\nomenclature{$\boldsymbol{\omega}_{\text{space}}$}{Hyperparameter combination search space}
\nomenclature{$\boldsymbol{\omega}$}{A hyperparameter combination}
\nomenclature{$\chi(\cdot)$}{Probability density function for hyperparameter configurations where \(y\) is better than the threshold \(y^*\)}
\nomenclature{$\gamma(\cdot)$}{Probability density function for hyperparameter configurations where \(y\) is worse than the threshold \(y^*\)}
\DeclareTextFontCommand{\texttimes}{\rmfamily\mdseries\upshape}
\usepackage{csquotes}

\usepackage{booktabs}
\usepackage{nomencl} 
\usepackage{tcolorbox} 
\usepackage{multicol} 
\usepackage{color}
\usepackage{threeparttable}
\usepackage{algorithm}
\usepackage{algpseudocode}
\usepackage{setspace}
\usepackage{amsthm}
\def\tsc#1{\csdef{#1}{\textsc{\lowercase{#1}}\xspace}}
\tsc{WGM}
\tsc{QE}
\tsc{EP}
\tsc{PMS}
\tsc{BEC}
\tsc{DE}
\usepackage{hyperref}
\usepackage{url}
\definecolor{lightblue}{RGB}{2,114,177}

\usepackage{graphicx}
\usepackage{booktabs}
\usepackage{multirow}
\usepackage{pifont}
\hypersetup{
    colorlinks=true,
    linkcolor=lightblue,
    filecolor=lightblue,      
    urlcolor=lightblue,
    citecolor=lightblue     
}

\begin{document}
\begin{sloppypar}
	\let\WriteBookmarks\relax
	\def\floatpagepagefraction{1}
	\def\textpagefraction{.001}
	\let\printorcid\relax
	\shorttitle{}
	\shortauthors{T. Cheng et~al.} 

	\title [mode = title]{A Tidal Current Speed Forecasting Model based on Multi-Periodicity Learning}



\author[1]{\textcolor[RGB]{0,0,1}{Tengfei Cheng}}

\author[1]{\textcolor[RGB]{0,0,1}{Yangdi Huang}}

\author[2]{\textcolor[RGB]{0,0,1}{Ling Xiao}}

\author[1]{\textcolor[RGB]{0,0,1}{Yunxuan Dong}}
\cormark[1]  
\ead{dyxiscool@outlook.com}

\address[1]{School of Computer, Electronics and Information, Guangxi University, Nanning, 530004, China}
\address[2]{College of Agriculture, Guangxi University, Nanning, 530004, China}
\cortext[cor1]{Corresponding author. \href{mailto:dyxiscool@outlook.com}{dyxiscool@outlook.com}} 

	\fntext[fn2]{This work was supported in part by the Department of Science and Technology of Guangxi Zhuang Autonomous Region (grant number 2024JJB170087), in part by the Department of Human Resources and Social Security of Guangxi Zhuang Autonomous Region (grant number 202401950).}


	\begin{abstract}
Tidal energy is one of the key components in increasing the penetration of renewable energy. High tidal energy penetration into the electrical grid depends on accurate tidal current speed forecasting. Model inaccuracies hinder forecast accuracy. Previous research primarily used physical models to forecast tidal current speed, yet tidal current variations influenced by the orbital periods of celestial bodies make accurate physical modeling challenging. Research on the multi-periodicity of tides is crucial for forecasting tidal current speed. We propose the Wavelet-Enhanced Convolutional Network to learn multi-periodicity. The framework embeds intra-period and inter-period variations of one-dimensional tidal current data into the rows and columns, respectively, of a two-dimensional tensor. Then, the two-dimensional variations of the sequence can be processed by convolutional kernels. We integrate a time-frequency analysis method into the framework to further address local periodic features. Additionally, to enhance the framework's stability, we optimize the framework's hyperparameters with the Tree-structured Parzen Estimator. The proposed framework captures multi-periodic dependencies in tidal current data. Numerical results show a 10-step average Mean Absolute Error of 0.025, with at least a 1.44\% error reduction compared to other baselines. Further ablation studies show a 1.4\% reduction in Mean Absolute Percentage Error on the data with artificially added periodic fluctuations.
	\end{abstract}


		

	\maketitle

	\section{Introduction}
 \subsection{Background}
The development of renewable energy can promote emission reduction \citep{CAGLAYAN2019113794}. Tidal energy is considered one of the most competitive forms of renewable energy \citep{OROURKE2014726}.

\begin{figure}
\small
\centering
\includegraphics[width=\columnwidth]{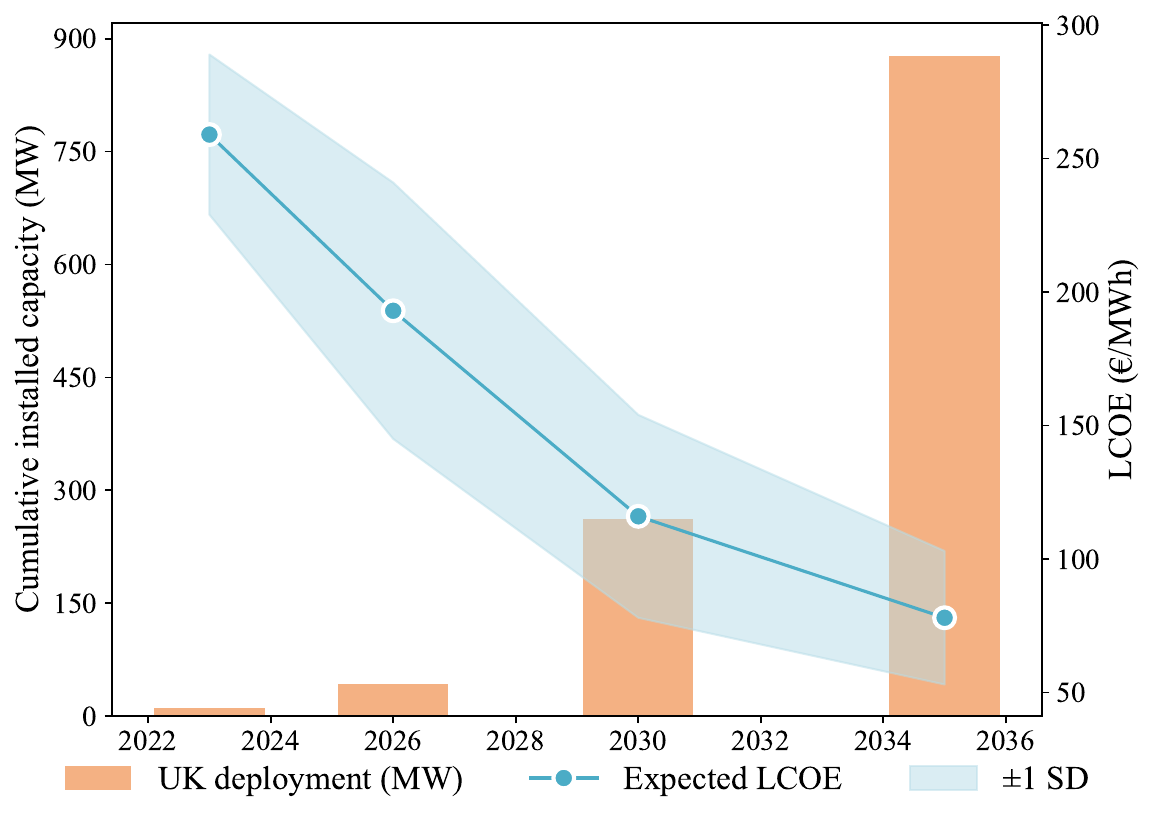}
\caption{Evaluation of Levelized Cost of Energy (LCOE) and Cumulative Installed Capacity in the UK. ±1 SD indicates data variation within one standard deviation around its expected value.} \label{1}
\end{figure}

Tidal energy attracts the attention of many coastal countries \citep{SHAO2023125476}, especially the UK. The potential of tidal energy lies in three aspects: cost-effectiveness, stability, and predictability. As tidal energy gains more attention, its cost-effectiveness is expected to improve, as shown in Figure \hyperref[1]{1}. Tidal energy may replace other renewable energies when the Levelized Cost of Energy (LCOE) falls below £49-55/MWh (expected by 2050) \citep{tiger2022cost}. Additionally, tidal energy offers high stability, which is crucial for the economic feasibility of any renewable energy project \citep{GRABBE20091898}. Therefore, the introduction of tidal energy can enhance grid stability. Moreover, tidal energy provides the advantage of predictability over large time scales. Consequently, tidal energy is an excellent choice as a base-load energy supply \citep{OROURKE2014726}.

Tidal current turbines generate electricity using the kinetic energy of tidal currents, like wind turbines use the kinetic energy of air \citep{YANG2024119603}. Therefore, fluctuations in Tidal Current Speed (TCS) reduce grid stability \citep{MONAHAN2023103596}. Enhancing the accuracy of short-term forecasting helps mitigate the impact of TCS fluctuations on grid stability, thereby increasing the penetration of tidal energy into the grid \citep{6665108}. Research \citep{wang2023wavelet} proposes to explore the statistical properties of real-life time series and design customized deep learning models for high-performance time series forecasting. The research on statistical properties of time series inspires us to explore the multi-periodicity within the TCS series. We increase the consideration of the statistical properties of TCS to improve the accuracy of TCSF.

\begin{table*}[htbp]
    \centering
    \begin{tcolorbox}[colback=white, colframe=black]
        \begin{multicols}{2}
            \printnomenclature
        \end{multicols}
    \end{tcolorbox}
\end{table*}

\subsection{Literature review}
Previous research focused on employing various models for TCSF. Existing tidal current forecasting models are divided into five categories: tidal harmonic models, physical models, statistical models, traditional machine learning models, and deep learning models. The summary of the advantages and disadvantages of the mentioned models are listed in Table \hyperref[11]{1}.

\begin{table}[htbp]
\centering
\begin{threeparttable}

    \begin{tabularx}{\columnwidth}{@{} >{\raggedright\arraybackslash}X c c c c c c @{}}
    \toprule

    \textbf{Forecasting model} & \tnote{\ding{172}} & \tnote{\ding{173}} & \tnote{\ding{174}} & \tnote{\ding{175}} & \tnote{\ding{176}} & \tnote{\ding{177}} \\
    \midrule
    Without Dependence on Specific Locations & {\color{red}\ding{55}} & {\color{red}\ding{55}} & {\color{green}\checkmark} & {\color{green}\checkmark} & {\color{green}\checkmark} & {\color{green}\checkmark} \\
    Considering local variations & {\color{green}\checkmark} & {\color{red}\ding{55}} & {\color{green}\checkmark} & {\color{red}\ding{55}} & {\color{red}\ding{55}} & {\color{green}\checkmark} \\
    Without Complex Feature Engineering & {\color{green}\checkmark} & {\color{green}\checkmark} & {\color{red}\ding{55}} & {\color{red}\ding{55}} & {\color{green}\checkmark} & {\color{green}\checkmark} \\
    Considering Multi-Periodicity & {\color{green}\checkmark} & {\color{red}\ding{55}} & {\color{red}\ding{55}} & {\color{red}\ding{55}} & {\color{red}\ding{55}} & {\color{green}\checkmark} \\
    \bottomrule
    \end{tabularx}

    \caption{The summary of the advantages and disadvantages of the mentioned models. The symbols \ding{172}-\ding{177} represent Tidal harmonic models, Physical models, Statistical models, Traditional machine learning models, Deep learning models, and Our models. respectively. Additionally, {\color{green}\checkmark} indicates the model has the corresponding feature, while {\color{red}\ding{55}} indicates it does not.}
    \label{11}

\end{threeparttable}
\end{table}

\paragraph{\texttimes{\textbf{Tidal harmonic models}}} In 1883, Sir G. H. Darwin proposed the harmonic analysis method \citep{Darwin_2009}. He stated, \enquote{The tidal oscillations of the ocean can be represented as the sum of several simple harmonic motions.} The harmonic analysis method was later developed by Doodson, who established the famous tidal harmonic component theory. Harmonic models rely on parameter estimation. Doodson continued this research and proposed estimating the parameters of the harmonic model with the least squares method \citep{doodson1957analysis}. Earlier technologies did not allow equipment to measure TCS. Research \citep{Fornerino_LeProvost_2015} applied harmonic models for TCSF latter. The harmonic model takes into account the multi-periodicity of TCS. However, the harmonic model cannot adapt to the local periodic changes of TCS. There were more models for TCSF as computing power increased.

\paragraph{\texttimes{\textbf{Physical models}}} The emergence of supercomputers brought advancements to physical models. Based on differential equations, physical models use two-dimensional or three-dimensional hydrodynamic models to make forecasts \citep{owen1980three}. Physical models can adapt to the non-stationarity of TCS. However, modeling with differential equations depends on physical knowledge. Additionally, solving these equations requires continuous measurement of boundary conditions. Therefore, data-driven models are needed for TCSF.

\paragraph{\texttimes{\textbf{Statistical models}}} Statistical models are suitable as a data-driven approach for TCSF. They avoid modeling based on specific TCS measurement locations. However, statistical models rely on strict stationarity assumptions. Therefore, statistical methods have some limitations and they have relatively high requirements in terms of data size, stability of time series, data distribution, and so on \citep{MA2020120159}.

\paragraph{\texttimes{\textbf{Machine learning models}}} With the rise of artificial intelligence technology, a large number of forecasting methods based on machine learning technology have emerged \citep{YANG2022119849}. Research \citep{5589321,sarkar2016machine, sarkar2018prediction} indicates that machine learning models outperform both statistical and harmonic models in predictive performance. However, machine learning models cannot determine data features. Feature extraction usually requires intuitive knowledge for quantification and expansion.

\paragraph{\texttimes{\textbf{Deep learning models}}} Deep learning models can provide accurate renewable energy generation forecasts \citep{Li2024}. Deep learning models are applied in the renewable energy field to reduce the complexity of feature engineering. Research \citep{9169644} used Long Short-Term Memory (LSTM) neural networks for tidal level forecasting. However, deep learning models are still underutilized in the tidal energy field. In other words, while deep learning methods have certain advantages, they are not designed for tidal current time series.

\subsection{Research gaps and proposed solution}

Existing research still has limitations. Tidal current time series demonstrate multi-periodicity \citep{qian2022tidal}. These time series include multiple periodic components, covering both astronomical cycles and local periods. Astronomical cycles include diurnal and semi-diurnal tides, among others. In addition to astronomical cycles, there are also local periodicities in tidal currents, manifested as local variations. For example, interactions between tides and waves cause local changes \citep{jay1999comparison}. Multiple overlapping and interacting cycles complicate the model of tidal current time series. Multi-periodicity presents an obstacle to TCSF. Secondly, existing research lacks deep learning models designed for tidal current time series. Therefore, using deep learning methods to consider the multi-periodicity in tidal current time series has sparked our research interest.

We are inspired by the TimesNet model proposed by Wu et al. \citep{wu2022timesnet}. The model focuses on multi-periodicity. The model uses the Fast Fourier Transform (FFT) to extract multiple periods from the time series. Based on the periods, the model transforms each 1D sequence into a 2D tensor. Then, the model uses visual backbone networks to capture the 2D variations in the time series. However, the framework lacks consideration for the local periodicity of tidal currents. Research \citep{ZHANG2022611} used time series decomposition methods to handle time series. The research inspired us: Discrete Wavelet Transform (DWT) can provide local information on signals in both time and frequency domains. We can handle the series at different scales to capture local periodicity. Additionally, the DWT module helps address local periodicity. The key challenge lies in integrating local periodicity into the learning of multi-periodicity. We propose a framework, Wavelet-Enhanced Convolutional Network (WECN), to address the aforementioned challenges. We use DWT to decompose the time series and measure sub-band amplitudes to identify dominant periods, feeding them into the predictor without reconstructing components. Finally, we consider the forecasting process to be an optimization problem. However, as the dimensionality of hyperparameters increases, the Gaussian model becomes increasingly complex. The Tree-structured Parzen Estimator (TPE) algorithm optimizes hyperparameters to handle the high dimensionality of the hyperparameter space \citep{KIM2024118010}.

\subsection{Novelty and contributions}
Overall, the contributions of this paper are as follows:
\begin{itemize}
\item We designed a time series forecasting framework for learning multi-periodicity. The framework outperforms existing models in terms of metrics.
\item We use DWT to account for local periodicity, aiming for better performance in multi-periodicity learning. Experiments validate the framework's performance.
\item We use TPE for hyperparameter optimization to improve forecasting accuracy. Ablation experiments demonstrate the effectiveness of the method.
\item  We design a novel framework for forecasting tidal current speed. The framework can contribute to advancements in the field of tidal energy.
\end{itemize}

\subsection{Paper organization}
This research is organized as follows: Section \ref{Section 2} shows the description of the multi-periodicity and the proposed framework.
Section \ref{Section 3} presents case studies on the tidal data.
Section \ref{Section 4} gives the conclusions and future works.

	\section{Methodology}
	\label{Section 2}
We construct a framework that considers the multi-periodicity of TCS. Section \ref{Section 2.1} shows the process of extracting periods using the improved wavelet. The periods extracted by DWT are used for the predictor. Section \ref{Section 2.2} shows the predictor used in this research. Section \ref{Section 2.3} shows the process of hyperparameter optimization.

Without loss of generality, a multivariate time series \(\textbf{\textit{X}}_{\text{1D}} = [\textit{\textbf{x}}^1, \textit{\textbf{x}}^2, \cdots, \textit{\textbf{x}}^n, \cdots,\textit{\textbf{x}}^N]\) is defined, where \(\textbf{\textit{X}}_{\text{1D}}\) is a two-dimensional array of shape \((T, N)\). The shape of \((T, N)\) represents \(N\) time series, each with \(T\) sampling points. \(\textbf{\textit{x}}^n\) denotes the \(n\)-th time series, \(\textbf{\textit{x}}^n = \{ \textit{x}^n(1),  \textit{x}^n(2), \ldots,  \textit{x}^n(T)\}\). 

 \subsection{Multi-scale time series analysis}
 \label{Section 2.1}
Multi-periodicity fundamentally arises from different inherent properties within the time series. The local periodicity of TCS is an intrinsic property in tidal current time series. We can analyze time series at multiple scales. We can identify information in both time and frequency domains. The approach is used to identify local periodicity. Therefore, we use the improved wavelet to extract periods for multi-scale time series analysis. We present the process of extracting periods. In this work, Discrete Wavelet Transform (DWT) is used solely for decomposition and amplitude-based measurement to select dominant scales; no inverse transform or signal reconstruction is performed.

DWT is first performed on the time series $\textbf{\textit{x}}^n$. The DWT uses a set of high-pass filters $h$ and low-pass filters $g$ to decompose the time series signal into different resolutions. As the decomposition levels increase, the frequency resolution of the original signal increases. At the $j$-th level, the outputs of the high-pass and low-pass filters are represented as detail coefficients $\textbf{\textit{cD}}^n_j$ and approximation coefficients $\textbf{\textit{cA}}^n_j$, respectively. The detail coefficients describe short-term changes in the series, while the approximation coefficients describe the long-term trends of the signal. The initial approximation is $\textbf{\textit{cA}}^n_0 = \textbf{\textit{x}}^n$. To obtain the $\textbf{\textit{cD}}^n_j$ and $\textbf{\textit{cA}}^n_j$ , the DWT operation at the $j$-th level is represented as:

\begin{equation}
\label{eq:1}
(\textbf{\textit{cD}}^n_j, \textbf{\textit{cA}}^n_j) = \text{DWT}(\textbf{\textit{cA}}^n_{j-1}) = (h * \textbf{\textit{cA}}^n_{j-1}, g * \textbf{\textit{cA}}^n_{j-1}),
\end{equation}where $j = 1, 2, ..., J$, $\text{DWT}(\cdot)$ denotes the execution of a single discrete wavelet transform, $*$ represents the convolution operation, and the selection of $h$ and $g$ depends on the wavelet basis $w$, a hyperparameter that needs optimization.

After the time series $\textbf{\textit{x}}^n$ is decomposed into $J$ levels, $J$ detail coefficients $[\textbf{\textit{cD}}^n_1,\textbf{\textit{ cD}}^n_2, ..., \textbf{\textit{cD}}^n_J]$ are output. At the $j$-th level, the detail coefficients $\textbf{\textit{cD}}^n_j$ has a length of $\frac{T}{2^j}$. For $\textbf{\textit{X}}_{\text{1D}}$, each series is decomposed into the same number of levels. The information from multiple series is first merged to obtain the average detail coefficients $\textbf{\textit{cD}}_j$, then the amplitude $\xi_j$ is calculated. $\xi_j$ can represent the intensity of signal change at that level, expressed as:

\begin{equation}
\label{eq:2}
\textbf{\textit{cD}}_j = \frac{1}{N} \sum_{n=1}^N \textbf{\textit{cD}}^n_j, \quad \xi_j = \|\textbf{\textit{cD}}_j\|^2,
\end{equation}where $j = 1, 2, \ldots, J$.

The detail layers with the largest amplitudes show the most significant frequency changes in the signal. Inspired by other research, to avoid meaningless noise from high frequencies, the top $k$ detail layers with the largest amplitudes are selected \citep{zhou2022fedformer}. $k$ is one of the hyperparameters to be optimized. If $J$ is less than $k$, all amplitudes are selected to form the amplitude array $\boldsymbol{\xi} = [\xi_1, \xi_2, \ldots, \xi_J]$, and $k$ is updated to $J$. If $J$ is greater than or equal to $k$, the top $k$ values are selected to form the amplitude array $\boldsymbol{\xi} = \text{top}(\xi_1, \xi_2, \ldots, \xi_J)$. Here, $\text{top}(\cdot)$ denotes the top $k$ elements selection to form a new array. Next, the index array $\psi = \text{index}(\xi)$ is obtained, where $\text{index}(\cdot)$ denotes the indices of the elements in the set. Finally, the frequency $f_i$ and period $p_i$ for the corresponding levels are calculated, expressed as:

\begin{equation}
f_i = \frac{F}{2^j}, \quad p_i = \left[\frac{1}{f_i}\right],
\label{eq:3}
\end{equation}where $i = 1, 2, \ldots, k$, $F$ denotes the sampling frequency of the time series data and $j \in \psi$.

Equation (\ref{eq:3}) results in the frequency $f_i$ and corresponding period $p_i$. To associate the amplitudes with their frequencies, the amplitudes are re-expressed as $\boldsymbol{A} = [A_{f_1}, \ldots, A_{f_k}]$.

\begin{figure}
\small
\centering
\includegraphics[width=\columnwidth]{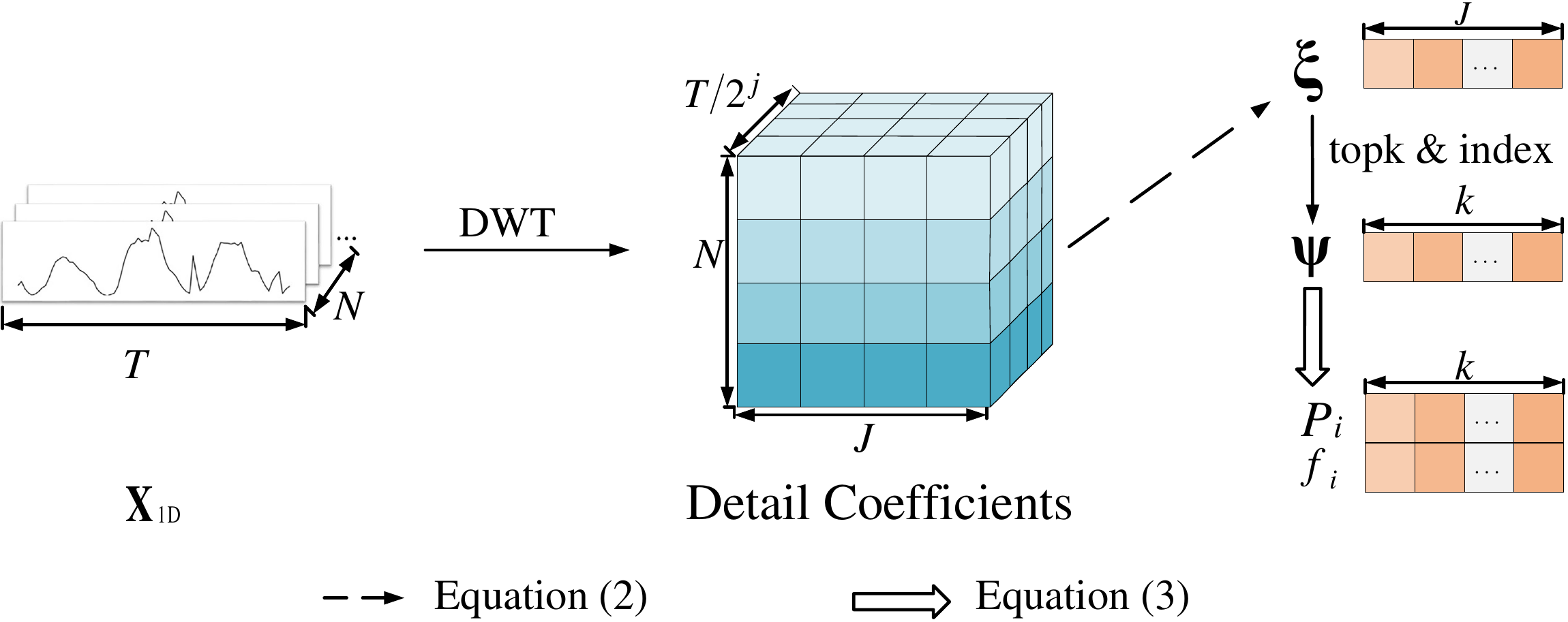}
\caption{Flowchart of the Period Extraction Module, illustrating the process of transforming multivariate time series inputs into the main period ($P$) and frequency ($f$).}
\label{2}
\end{figure}

\subsection{Tidal current speed predictor}
 \label{Section 2.2}
We use WECN to forecast after we obtain the periodic information. Existing methods have limitations in capturing multi-periodicity. Existing methods are either limited by one-dimensional input sequences, making it difficult to capture multi-periodicity, or require complex feature engineering \citep{jia2024witran}. This research employs the WECN framework to model tidal current time series, as it can fully utilize periodic information for forecasting. The WECN framework uses the TimesNet model proposed by Wu et al. as the infrastructure of the predictor. Wu et al. \citep{wu2022timesnet} observed that the time point variation for each period is influenced by the temporal patterns of its neighboring region and changes in its adjacent periods. Therefore, the two-dimensional changes in the time series can be represented as intra-period and inter-period variations.

\begin{figure*}
 \centering
        \includegraphics[width=0.95\linewidth]{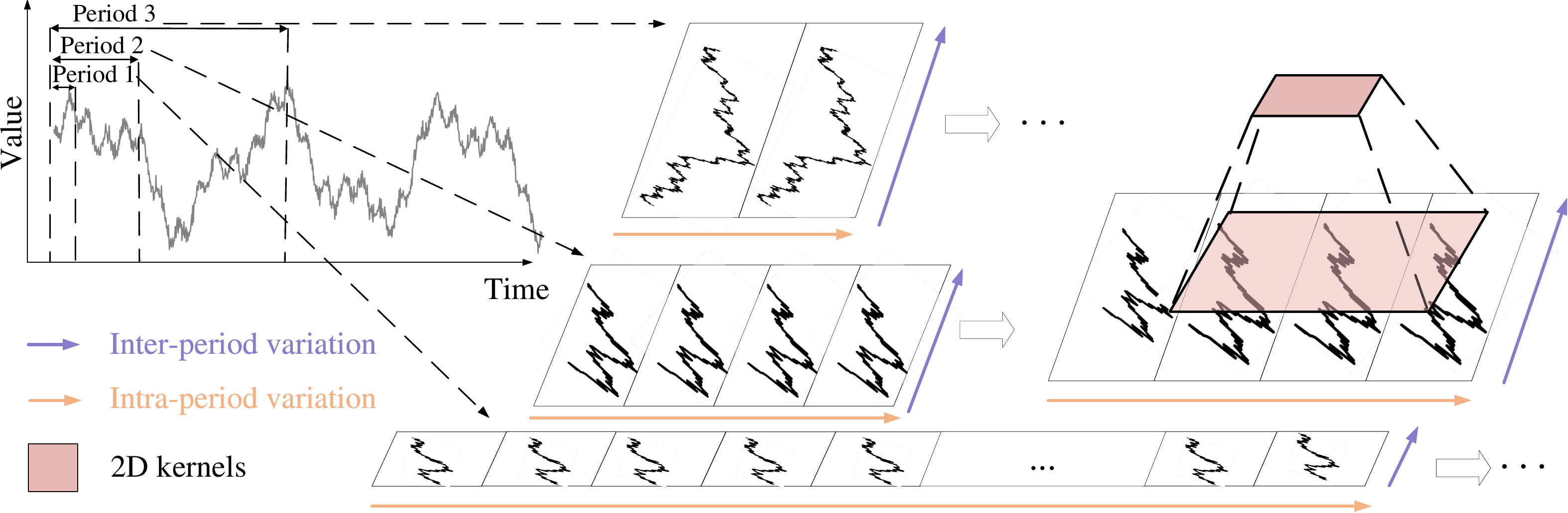}
        \caption{The WECN captures intra-period and inter-period variations and uses 2D convolutional kernels to process both types of variations. The time series case in the figure is selected from the used tidal current speed dataset. } 
        \label{3} 
    \end{figure*}

To uniformly represent intra-period and inter-period variations, the original one-dimensional time series is transformed into a set of two-dimensional tensors based on multiple periods. The WECN framework uses visual backbone networks to capture the two-dimensional changes in the time series and achieve a unified model of intra-period and inter-period variations, as shown in Figure \hyperref[3]{3}.

First, the original input $\textbf{\textit{X}}_{\text{1D}}$ is projected into deep features through an embedding layer, represented as:
\begin{equation}
\textbf{\textit{X}}_{\text{1D}}^0 = \text{Embed}(\textbf{\textit{X}}_{\text{1D}}),
\end{equation}
where $\text{Embed}(\cdot)$ represents the embedding layer. $\textbf{\textit{X}}_{\text{1D}}^0$ is a two-dimensional array of shape $(T, d)$, representing the zeroth layer features in TimesNet. Here, $d$ is a hyperparameter to be optimized, representing the feature space dimension.

\begin{figure}
\small
\centering
\includegraphics[width=\columnwidth]{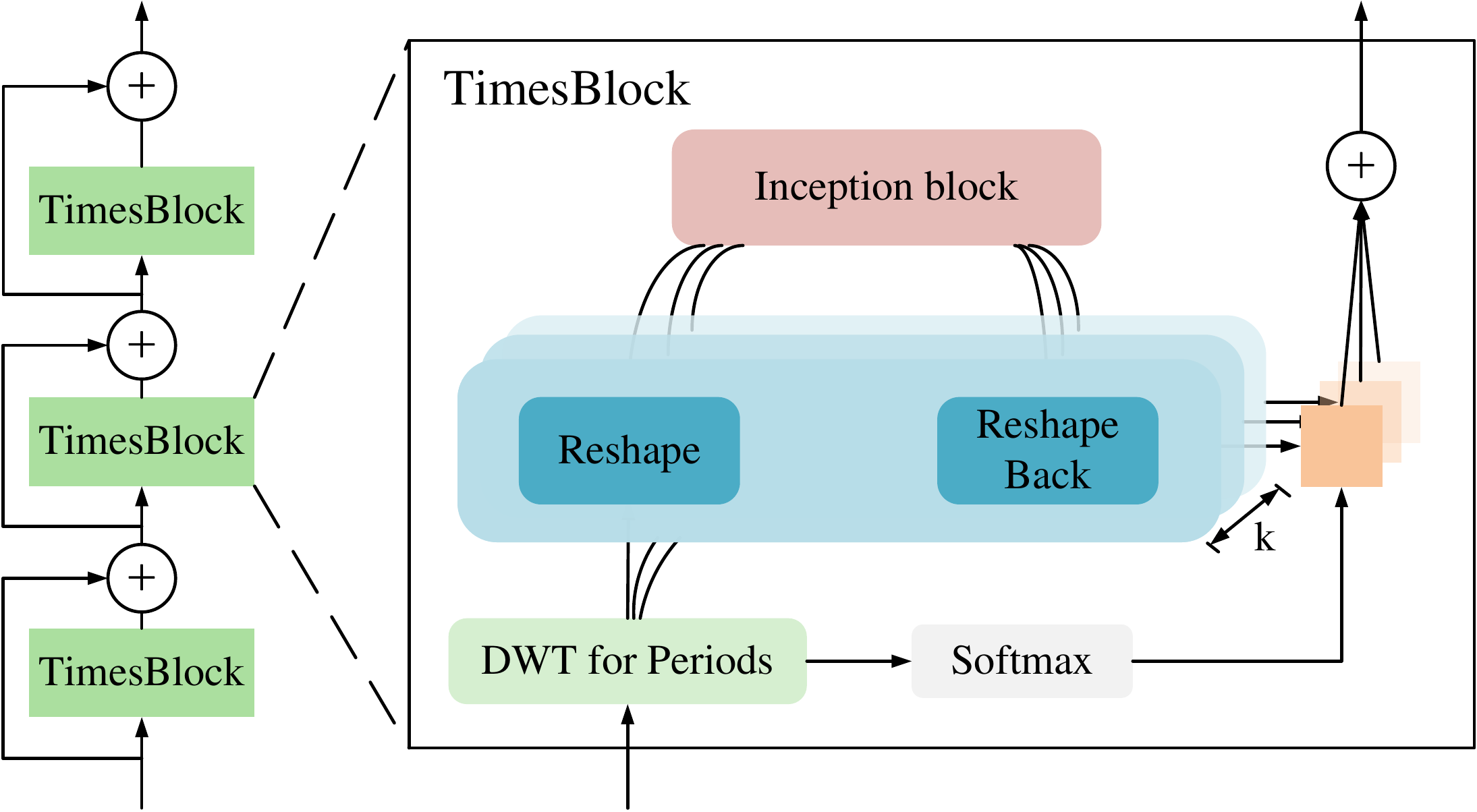}
\caption{Overall architecture of WECN.} \label{4}
\end{figure}

The adjacent TimesBlock structures are connected using the Residual structure \citep{he2016deep}. The WECN framework is illustrated in Figure \ref{4}.

To obtain the features of the $l$-th layer in TimesNet, the input is $\textbf{\textit{X}}_{\text{1D}}^{l-1}$, represented as:
\begin{equation}
\textbf{\textit{X}}_{\text{1D}}^l = \text{TimesBlock}(\textbf{\textit{X}}_{\text{1D}}^{l-1}) + \textbf{\textit{X}}_{\text{1D}}^{l-1}.
\end{equation} $\text{TimesBlock}(\cdot)$ performs three steps: ascending dimension, capturing temporal 2D-variations, and adaptive aggregation.

\paragraph{\texttimes{\textbf{Ascending dimension}}}
The input of the $l$-th layer is $\textbf{\textit{X}}_{\text{1D}}^{l-1}$, and the calculations in Equations (\ref{eq:1}) - (\ref{eq:3}) are encapsulated in the function Period $(\cdot)$, expressed as:
\begin{equation}
\textbf{\textit{A}}^{l-1}, \{f_1, \cdots, f_k\}, \{p_1, \cdots, p_k\} = \text{Period}(\textbf{\textit{X}}_{\text{1D}}^{l-1}),
\end{equation}where $\textbf{\textit{A}}^{l-1}=\{A_{f_1}^{l-1}, \cdots, A_{f_k}^{l-1}\}$.

Based on the frequency set $\{f_1, \cdots, f_k\}$ and the corresponding period lengths $\{p_1, \cdots, p_k\}$ selected by DWT, 
the required number of periods is computed as $q_i = \left[\frac{T}{p_i}\right]$, 
and the 1D time series $\textbf{\textit{X}}_{\text{1D}}$ is transformed into multiple 2D tensors through the following equation:
\begin{equation}
\textbf{\textit{X}}_{\text{2D}}^{l, i} = \text{Reshape}_{p_i, q_i}(\text{Padding}(\textbf{\textit{X}}_{\text{1D}}^{l-1})),
\end{equation}where $i = 1, 2, \cdots, k$, $\textbf{\textit{X}}_{\text{1D}}^{l-1}$ is a two-dimensional array of shape $(T, d)$. $\text{Padding}(\cdot)$ applies zero-padding to the sequence, extending each sequence length from $T$ to $(p_i \times q_i)$. $\text{Reshape}_{p_i, q_i}(\cdot)$ reshapes the sequence into a 2D tensor, resulting in $\textbf{\textit{X}}_{\text{2D}}^{l, i}$ as a two-dimensional array of shape $(p_i \times q_i, d)$.

The columns and rows of $\textbf{\textit{X}}_{\text{2D}}^{l, i}$ represent intra-period and inter-period variations corresponding to the period length. This arrangement allows 2D kernels to handle 2D-time variations easily.

\paragraph{\texttimes{\textbf{Capturing temporal 2D-variations}}}
2D-time series features can be extracted using convolution operations. Furthermore, the model uses the Inception V1 structure from GoogLeNet \citep{7298594}. For the 2D tensor, the convolution process can be expressed as:
\begin{equation}\label{9}
\hat{\mathbf{X}}^{l,i}_{\text{2D}} =\operatorname{Inception}\left(\mathbf{X}^{l,i}_{\text{2D}}\right),\ i =1, 2, ..., k. \\
\end{equation}

\paragraph{\texttimes{\textbf{Adaptive aggregation}}}

The data is first transferred from 2D to 1D space with the same period size. The Trunc$(\cdot)$ function truncates the padded sequence of length $(p_i \times q_i)$ to its original length $T$, expressed as:
\begin{equation}
\hat{\mathbf{X}}^{l,i}_{\text{1D}} = \text{Trunc}\left(\text{Reshape}_{1,(p_{i} \times q_{i})}\left(\hat{\mathbf{X}}^{l,i}_{\text{2D}}\right)\right),\ i =1, 2, ..., k.
\end{equation}

Next, we need to merge the $k$ different 1D representations $\{\hat{\mathbf{X}}^{l,1}_{\text{1D}},\cdots,\hat{\mathbf{X}}^{l,k}_{\text{1D}}\}$  for the subsequent layer. The amplitude $A$ reflects the relative importance of the selected frequencies and periods, which correlates with the significance of each transformed 2D tensor \citep{NEURIPS2021_bcc0d400}. Therefore, the Softmax function calculates the weights. After this, the 1D representations based on amplitude are aggregated.

\begin{equation}\label{equ:aggregation}
  \begin{split}
  \hat{\mathbf{A}}^{l-1}_{f_{1}},\cdots,\hat{\mathbf{A}}^{l-1}_{f_{k}} & = \mathrm{Softmax}\left(\mathbf{A}^{l-1}_{f_{1}},\cdots,\mathbf{A}^{l-1}_{f_{k}}\right), \\
  {\mathbf{X}}_{\text{1D}}^{l} & =\sum_{i=1}^{k}\hat{\mathbf{A}}_{f_{i}}^{l-1}\times \hat{\mathbf{X}}^{l,i}_{\text{1D}}.\\
  \end{split}
\end{equation}

\begin{figure}
\small
\centering
\includegraphics[width=\columnwidth]{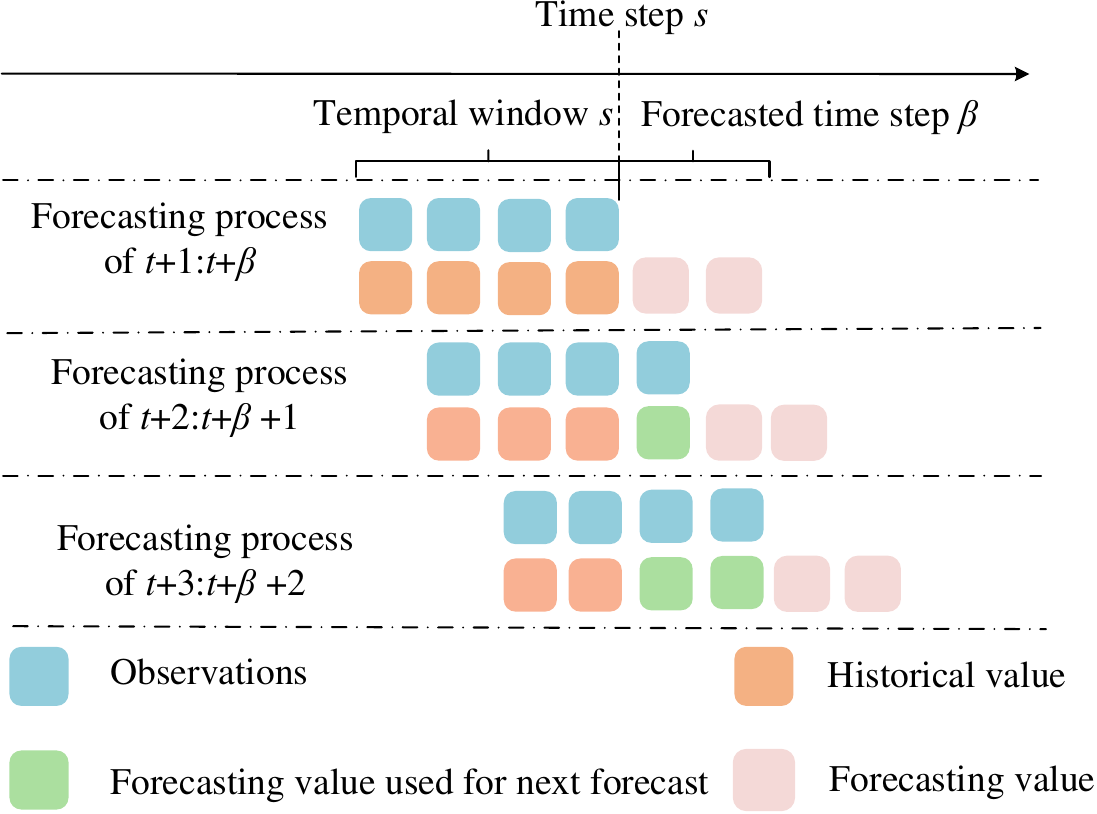}
\caption{The multi-step forecasting process of WECN.} \label{5}
\end{figure}

$\textbf{\textit{X}}_{\text{1D}}$ is processed through each TimesBlock with residual connections. Then, the features from the final layer are projected into the output space via a fully connected layer for multi-step forecasting. Single-step forecasting is sometimes insufficient to ensure the reliability and controllability of tidal energy systems. Therefore, this paper adopts and discusses the multi-step forecasting mechanism, providing more information for future decision-making \citep{du2019novel}. In practical tidal energy systems, forecasts are typically updated in an online and rolling manner as new observations become available. Recursive forecasting enables low-latency step-by-step prediction while supporting flexible extension of the forecasting horizon. Under such real-time operational constraints, we use a recursive forecasting strategy for multi-step forecasting, where the current output forecasting serves as the input for the next step. The multi-step forecasting process can be seen in Figure \ref{5}. The multi-step forecast results belong to $\mathbb{R}^{T \times N \times \beta}$, where $\beta$ denotes the output length. The $g$-th step forecasting value for the $t$-th sampling point in the $n$-th sequence can be represented as $\hat{x}^{n}(t+g)$, where $g = 1, 2, ..., \beta$. 

\subsection{Interpretation of the tidal receptive field}

Tidal current speed contains (i) phase evolution within one tidal cycle and (ii) cycle-to-cycle modulation across consecutive cycles.
Equation (7) reshapes $\textbf{\textit{X}}_{\text{1D}}^{l-1}$ into $\textbf{\textit{X}}_{\text{2D}}^{l,i}$ so that the 2D receptive
field used in Equation (9) aligns with these two physical structures: the row axis corresponds to intra-period (within-cycle) phase
neighborhoods, and the column axis corresponds to inter-period (across-cycle) neighborhoods.

We define an index mapping 
from the 2D coordinates of $\textbf{\textit{X}}_{\text{2D}}^{l,i}$ to the original 1D position:
\begin{equation}
\tau(r,c)=(c-1)p_i+r,\qquad r\in\{1,\ldots,p_i\},\ c\in\{1,\ldots,q_i\}.
\end{equation}
Thus, moving along $r$ captures local phase-shape changes within one cycle, while moving along $c$ compares similar phases across
neighboring cycles.

If a kernel in Equation (9) covers an $a\times b$ patch on $\textbf{\textit{X}}_{\text{2D}}^{l,i}$, then the corresponding maximum
span on the original 1D index is
\begin{equation}
\Delta \tau_{\max}=(b-1)p_i+(a-1),
\end{equation}
and the physical time span equals $\Delta \tau_{\max}\Delta_s$, where $\Delta_s$ is the sampling interval. Therefore, $a$ controls
the within-cycle (phase-neighborhood) coverage and $b$ controls the across-cycle coverage for capturing cycle-to-cycle modulation
and slow drift.

\subsection{The process of hyperparameter optimization}
 \label{Section 2.3}
 We recognize the importance of selecting optimal hyperparameters. To achieve this, we use the Tree-structured Parzen Estimator (TPE). TPE is a variant of Bayesian optimization suitable for high-dimensional and discrete parameter spaces. Unlike traditional Bayesian methods that rely on Gaussian processes, TPE improves efficiency by constructing probability density functions in the hyperparameter search space using Parzen window density estimation \citep{ALISALAMAI2023118658}. Next, we transform the forecasting problem into an optimization process, with the design of the objective function discussed in Section \ref{Section 2.3.1} and the application of TPE in Section \ref{Section 2.3.2}.

 \subsubsection{\texttimes{\textbf{Objection function}}}
 
 \label{Section 2.3.1}
 Typically, multi-step time series forecasting aims to minimize the Mean Squared Error (MSE). We define the following equation as the optimization objective:

\begin{equation}\label{e}
  \begin{aligned}
    &  \min_{\boldsymbol{\omega}} \ y, \\
     \text{s.t.} \ &y = \frac{1}{\beta} \sum_{g=1}^{\beta} \left( \frac{1}{NT} \sum_{t=1}^{T} \sum_{n=1}^{N} \left(\hat{x}^{n}(t+g, \boldsymbol{\omega})- x^{n}(t+g, \boldsymbol{\omega})\right)^2 \right), \\
    & \boldsymbol{\omega} \in \boldsymbol{\omega}_{\text{space}},
  \end{aligned}
\end{equation}
where $\boldsymbol{\omega}$ represents a combination of hyperparameters within the hyperparameter search space, $\hat{x}^{n}(t+g, \boldsymbol{\omega})$ represents the forecasted value, $x^{n}(t+g, \boldsymbol{\omega})$ represents the observed value. $\boldsymbol{\omega}_{\text{space}}$ denotes the hyperparameter search space, and $y$ is the value of the loss function.

\subsubsection{\texttimes{\textbf{Implementation of TPE}}}

 \label{Section 2.3.2}
TPE continuously searches for the optimal solution of the objective function in a tree-structured parameter space. Here, $\boldsymbol{\omega}$ represents any hyperparameter combination within the hyperparameter search space $\boldsymbol{\omega}_{\text{space}}$, and $y$ is the loss function value under that hyperparameter combination. The optimization process of TPE is illustrated in Algorithm \hyperref[111]{1}, which involves four main steps:

\paragraph{\texttimes{\textbf{Initial distribution definition}}} Perform a random search in the initial iteration phase to help establish the initial probability distribution.

\paragraph{\texttimes{\textbf{Observation splitting}}} Divide the hyperparameter combinations into two groups based on their performance metrics. The first group includes ``good'' configurations, denoted as $\chi(\boldsymbol{\omega})$, and the second group includes ``bad'' configurations, denoted as $\gamma(\boldsymbol{\omega})$. The performance threshold $y^*$ distinguishes two groups. The groups are updated after evaluating the performance of new hyperparameter combinations.

\paragraph{\texttimes{\textbf{Probability modeling}}} Use Parzen window estimation to build probability density functions for the configurations $\chi(\boldsymbol{\omega})$ and $\gamma(\boldsymbol{\omega})$. Specifically, $p(\boldsymbol{\omega} \mid y < y^*)$ is used for ``good'' configurations, and $p(\boldsymbol{\omega} \mid y \geq y^*)$ is used for ``bad'' configurations.

\paragraph{\texttimes{\textbf{Sampling strategy}}} In each iteration, select new hyperparameter combinations for evaluation based on the current probability density distributions. The selection strategy maximizes the Expected Improvement (EI), representing the likelihood that a new hyperparameter configuration will further reduce the loss, given the current best loss function value. It is calculated as follows:
\begin{equation}
\operatorname{EI}(\boldsymbol{\omega}) = \frac{\chi(\boldsymbol{\omega})} {\gamma(\boldsymbol{\omega})}. 
\end{equation}The hyperparameters that maximize EI are selected as the next set of search values, and the process returns to step 2 for iteration.

\paragraph{\texttimes{\textbf{Output}}} When the stopping criteria for iteration are met, the best hyperparameter combination $\boldsymbol{\omega}_{best}$ is output. 
\begin{algorithm}
\label{111}
\caption{Tree-structured Parzen Estimator}
\begin{algorithmic}[1]
\State Initial distribution definition;
\While{\underline{the stopping criteria for iteration are not reached}}
    \State Construct models $ \chi(\boldsymbol{\omega})$ and $\gamma(\boldsymbol{\omega})$;
    \State Select new hyperparameters $\boldsymbol{\omega}$ to maximize $EI(\boldsymbol{\omega}) = \frac{ \chi(\boldsymbol{\omega})}{\gamma(\boldsymbol{\omega})}$;
    \State Evaluate performance of $\boldsymbol{\omega}$ to obtain $y$;
    \State Update configuration groups;
\EndWhile
\State \textbf{end}
\State Return $\boldsymbol{\omega}_{best}$;
\end{algorithmic}
\end{algorithm}

	\section{Experiment}
\label{Section 3}

\subsection{Experimental setup}

All experiments were conducted on a GeForce RTX 4090 GPU. We used a Linux operating system and the PyTorch deep learning framework under Python 3.8. To ensure the reproducibility of our experimental results, our core Python packages were aligned with the Time Series codebase, such as numpy==1.23.5 and pandas==1.5.3. The measurements are sampled at 1\,Hz, and one time step equals 1\,s throughout the paper. In the optimization experiments, to balance exploration of the hyperparameter space and computational efficiency, we set the stopping criterion for optimization to a maximum of 100 evaluations. Among these evaluations, the hyperparameter configuration that achieves the lowest validation loss is selected as the optimal setup for our final model.

Time series forecasting is highly susceptible to look-ahead bias, particularly when future information leaks into preprocessing, model training, or model selection. In this work, we strictly follow a modeling protocol:
\begin{itemize}
  \item The training, validation, and testing sets are split in chronological order, ensuring that the validation and test segments occur strictly after the training segment;
  \item No future samples are used during preprocessing, i.e., any parameters or statistics required by preprocessing are estimated using the training segment only and then applied unchanged to the validation and test sets;
  \item Training samples are constructed using causal input windows, such that each target value is predicted solely from historical observations.
\end{itemize}
This protocol ensures that the reported performance faithfully reflects realistic forward-in-time forecasting.

\subsection{Datasets}
The Orkney Islands in the United Kingdom, located in the North Atlantic north of Scotland, are renowned for their abundant tidal energy resources and are known as an ``ecological paradise'' in the marine energy sector. Their unique geographical location provides real marine environmental conditions for the tidal energy test site at the European Marine Energy Centre, including various factors such as tides, ocean currents, and water depth. These data are crucial for verifying the performance of tidal energy technology in actual environments.

\begin{figure*} 
\centering
\includegraphics[width=17cm]{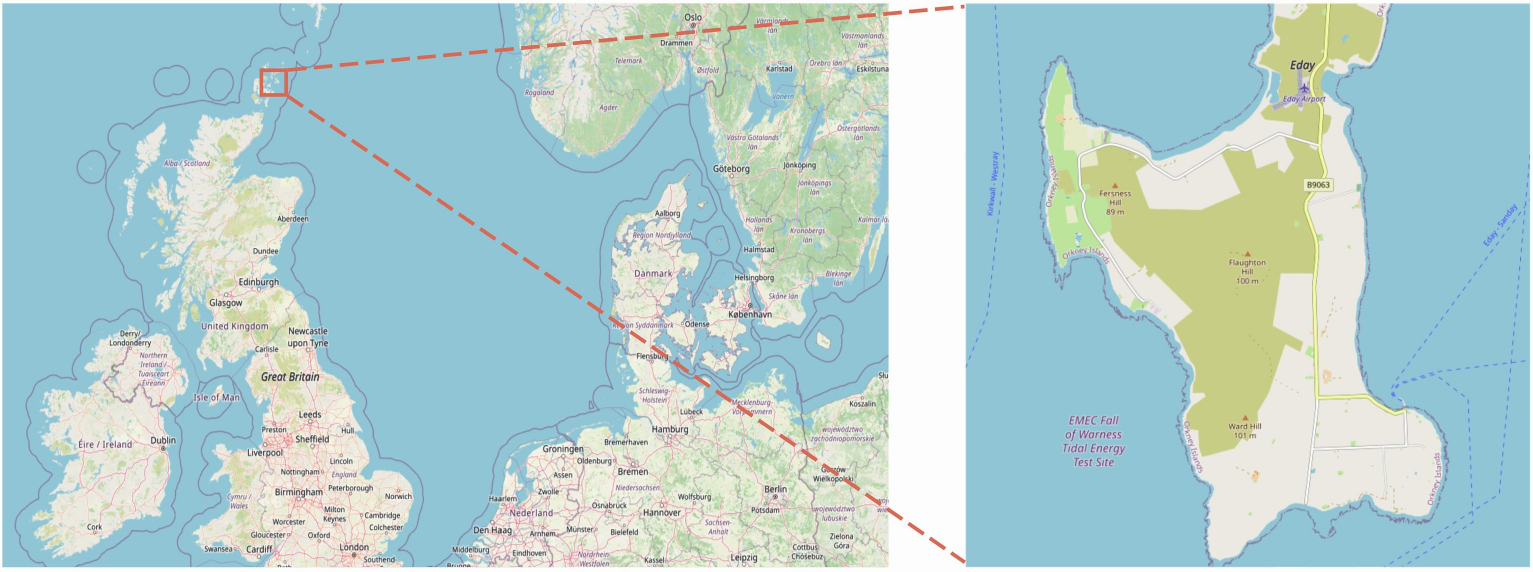}
\caption{Location of the tidal energy test site near Fall of Warness, UK.} \label{6}
\end{figure*}

The tidal data used in this paper were obtained from \href{https://datashare.ed.ac.uk/handle/10283/4422}{ReDAPT project}, originating from a tidal energy test site near the Fall of Warness on one of the Orkney Islands. The location can be seen in Figure \ref{6}. The TCS data were collected from four-beam Acoustic Doppler Current Profilers instruments deployed on the seabed gravity mooring frames at the tidal energy test site. 

The dataset was recorded continuously over a seven-day period with a high-resolution sampling interval of 1~s, allowing for detailed temporal analysis. The 1~s sampling interval is particularly important for capturing fine-grained dynamics and turbulent fluctuations inherent in the system.

 The dataset is deliberately designed to support an evaluation of model generalization under distributional variations. Specifically, the velocity components along the X, Y, and Z directions are treated as independent forecasting targets to capture direction-dependent tidal flow dynamics. In addition, multi-level forecasting is conducted at three different depths (22,m, 23,m, and 24,m below sea level), introducing hydrodynamic variations induced by depth differences. These settings expose the model to multiple related but non-identical data distributions, enabling an assessment of whether the learned representations generalize beyond a single channel or a single depth. This design allows the proposed method to be evaluated across multiple velocity components and depths within a consistent physical environment, providing a controlled yet non-trivial test of generalization under limited but realistic high-frequency tidal measurements.

The original data are stored using .mat file format. We first perform preprocessing. Due to malfunctions in reading instruments or interruptions caused by sudden movements near recording sensors, the data contain missing values and outliers. For non-consecutive missing values, gaps are filled using the average of neighboring time points. This approach provides a reasonable local estimate while preserving the temporal continuity of the series without introducing significant bias. For consecutive missing segments, the data are directly replaced with a default value of 0.01. This small constant is used to ensure numerical stability and consistency during processing, to clearly distinguish imputed values from true zeros, and to minimize their impact on the overall statistical properties of the dataset. \citep{Chavat2022ECDUY}. One method to remove outliers is to pre-filter high-frequency content in the data \citep{5589321}. This research uses an adaptive filter to remove high-frequency noise and erroneous data from the dataset. Table \ref{22} shows statistics of the data at different water depths.

\begin{table}[htbp]
\centering

\resizebox{\columnwidth}{!}{
\begin{tabular}{@{}p{0.2\columnwidth}p{0.2\columnwidth}p{0.2\columnwidth}p{0.2\columnwidth}p{0.2\columnwidth}@{}}
\toprule
\textbf{BSL} & \textbf{Direction} & \textbf{Mean} & \textbf{Median} & \textbf{Std}  \\
\midrule
\multirow{3}{*}{22m}
& X & -0.0633 & -0.1798 & 1.3070   \\
& Y & 0.0844 & -0.0716 & 1.3545  \\
& Z & 0.0074 & 0.0072 & 0.0699  \\
\midrule
\multirow{3}{*}{23m}
& X & -0.0479 & -0.1769 & 1.3272   \\
& Y & 0.0852 & -0.0731 & 1.3656   \\
& Z & 0.0075 & 0.0072 & 0.0684   \\
\midrule
\multirow{3}{*}{24m}
& X & -0.0343 & -0.1754 & 1.3455   \\
& Y & 0.0865 & -0.0804 & 1.3771  \\
& Z & 0.0077 & 0.0069 & 0.0671 \\
\bottomrule
\end{tabular}
}
\caption{BSL stands for Below Sea Level. Mean denotes the mean values of the time series, Median denotes the intermediate value of the time series, Std is the standard deviation of the time series. The unit of tidal current velocity is meters per second (m/s).
}
\label{22}
\end{table}

We split the data into training, testing, and validation sets with proportions of 70\%, 20\%, and 10\%, respectively. The validation set is designed to adjust the hyperparameters. We select TCS for a case study to provide a practical dataset and method application example.

\subsection{Evaluation metrics}
This subsection describes the performance metrics adopted to evaluate the forecasting capabilities of the proposed framework. Specifically, Mean Absolute Error (MAE), Mean Square Error (MSE), Mean Absolute Percentage Error (MAPE) are used as follows:

\begin{equation}
    \text{MAE} = \frac{1}{T}\sum_{t=1}^{T}|\hat{x}^{n}(t) - x^{n}(t)|
\end{equation}
\begin{equation}
    \text{MSE} = \frac{1}{T}\sum_{t=1}^{T}(\hat{x}^{n}(t) - x^{n}(t))^2
\end{equation}

\begin{equation}
    \text{MAPE} = \frac{1}{T}\sum_{t=1}^{T}\left|\frac{\hat{x}^{n}(t) - x^{n}(t)}{x^{n}(t)}\right| \times 100\%
\end{equation}

where $x^{n}(t)$ represents the observations in the datasets, $\hat{x}^{n}(t)$ is the forecasting
results of the model, and $T$ denotes the number of
forecasting times. 

This model's training utilizes a typical supervised learning framework. The metrics for the three directions at the same depth are averaged. The objective function of the optimization method is to minimize the average value of MSE.

\subsection{Performance comparisons of different models}
We conduct a comparative analysis against existing methods. 

\paragraph{\texttimes{\textbf{Baselines}}}
We select the following models as baselines: 
XGBoost \citep{chen2016xgboost}, 
LSTM \citep{hochreiter1997long}, 
GRU \citep{chung2014empirical}, 
Reformer \citep{Kitaev2020Reformer}, 
Autoformer \citep{wu2021autoformer}, 
DLinear \citep{Zeng_Chen_Zhang_Xu_2023}, 
FEDformer \citep{pmlr-v162-zhou22g} 
and FiLM \citep{NEURIPS2022_524ef58c}. 
The comparison models' hyperparameter design refer to the work \citep{wang2024tssurvey}.

\subsection{Performance comparisons of different models}
We conduct a comparative analysis against existing methods.

\begin{table*}[p]
\caption{Metric values of different models for multi-step forecasting.}
\label{7} 
\centering

\begin{minipage}{\linewidth}
\centering
\textbf{(a) 10-step forecasting}

\vspace{0.5em}

\resizebox{\linewidth}{!}{
\begin{tabular}{cccccccccc}
\toprule
\multicolumn{1}{c}{\multirow{2}{*}{BSL}} &
\multicolumn{3}{c}{22m} &
\multicolumn{3}{c}{23m} &
\multicolumn{3}{c}{24m} \\
\cmidrule(lr){2-4} \cmidrule(lr){5-7} \cmidrule(lr){8-10}
Metrics & MAE & MSE & MAPE (\%) & MAE & MSE & MAPE (\%) & MAE & MSE & MAPE (\%) \\
\midrule
XGBoost & 0.2489 & 0.1137 & 485.51 & 0.2504 & 0.1150 & 507.83 & 0.2517 & 0.1160 & 505.37 \\
LSTM & 0.1014 & 0.0210 & 335.15 & 0.0926 & 0.0187 & 259.02 & 0.1025 & 0.0223 & 297.83 \\
GRU & 0.2530 & 0.1001 & 1242.34 & 0.2666 & 0.1090 & 1154.99 & 0.2589 & 0.1072 & 1154.47 \\
Reformer & 0.0662 & 0.0100 & 47.37 & 0.0646 & 0.0103 & 62.00 & 0.0662 & 0.0107 & 99.90 \\
Autoformer & 0.1601 & 0.0891 & 176.06 & 0.1961 & 0.1348 & 236.62 & 0.1699 & 0.1060 & 297.15 \\
DLinear & \textcolor[rgb]{0.804,0.498,0.196}{0.0347} & 0.0062 & \textcolor[rgb]{0.804,0.498,0.196}{35.17}
        & \textcolor[rgb]{0.804,0.498,0.196}{0.0344} & 0.0062 & 41.87
        & \textcolor[rgb]{0.804,0.498,0.196}{0.0340} & 0.0061 & \textcolor[rgb]{0.804,0.498,0.196}{62.68} \\
FEDformer & 0.0518 & \textcolor[rgb]{0.804,0.498,0.196}{0.0052} & 50.68
          & 0.0429 & \textcolor[rgb]{0.804,0.498,0.196}{0.0037} & \textcolor[rgb]{0.804,0.498,0.196}{40.91}
          & 0.0448 & \textcolor[rgb]{0.804,0.498,0.196}{0.0039} & 164.95 \\
FiLM & \textcolor[rgb]{0.753,0.753,0.753}{0.0257} & \textcolor[rgb]{0.753,0.753,0.753}{0.0033} & \textcolor[rgb]{1.0,0.874,0.0}{29.63}
     & \textcolor[rgb]{0.753,0.753,0.753}{0.0255} & \textcolor[rgb]{0.753,0.753,0.753}{0.0033} & \textcolor[rgb]{0.753,0.753,0.753}{29.15}
     & \textcolor[rgb]{1.0,0.874,0.0}{0.0251} & \textcolor[rgb]{0.753,0.753,0.753}{0.0032} & \textcolor[rgb]{1.0,0.874,0.0}{47.68} \\
\textbf{WECN (Ours)} &
\textcolor[rgb]{1.0,0.874,0.0}{0.0254} & \textcolor[rgb]{1.0,0.874,0.0}{0.0025} & \textcolor[rgb]{0.753,0.753,0.753}{30.14}
& \textcolor[rgb]{1.0,0.874,0.0}{0.0245} & \textcolor[rgb]{1.0,0.874,0.0}{0.0029} & \textcolor[rgb]{1.0,0.874,0.0}{24.74}
& \textcolor[rgb]{0.753,0.753,0.753}{0.0253} & \textcolor[rgb]{1.0,0.874,0.0}{0.0030} & \textcolor[rgb]{0.753,0.753,0.753}{53.83} \\
\bottomrule
\end{tabular}}
\end{minipage}

\vspace{1.2em}

\begin{minipage}{\linewidth}
\centering
\textbf{(b) 30-step forecasting}

\vspace{0.5em}
\resizebox{\linewidth}{!}{
       \begin{tabular}{cccccccccc}
            \toprule
            \multicolumn{1}{c}{\multirow{2}{*}{BSL}} &
            \multicolumn{3}{c}{22m} &
            \multicolumn{3}{c}{23m} &
            \multicolumn{3}{c}{24m} \\
            \cmidrule(lr){2-4} \cmidrule(lr){5-7} \cmidrule(lr){8-10}
            \multicolumn{1}{c}{Metrics} & MAE & MSE & MAPE (\%) & MAE & MSE & MAPE (\%) & MAE & MSE & MAPE (\%) \\
            \midrule
            XGBoost \citep{chen2016xgboost} & 0.2546 & 0.1190 & 923.65 & 0.2562 & 0.1203 & 899.87 & 0.2573 & 0.1212 & 922.76 \\
            LSTM \citep{hochreiter1997long} & 0.2096 & 0.1929 & 979.73 & 0.2398 & 0.2147 & 979.39 & 0.2140 & 0.2987 & 1009.88 \\
            GRU \citep{chung2014empirical} & 0.2877 & 0.1359 & 1452.25 & 0.2789 & 0.1283 & 1330.46 & 0.2820 & 0.1327 & 1349.83 \\
            Reformer \citep{Kitaev2020Reformer} & 0.1659 & \textcolor[rgb]{0.804, 0.498, 0.196}{0.1031} & \textcolor[rgb]{0.753, 0.753, 0.753}{120.16} & 0.1709 & 0.1095 & 147.18 & 0.1710 & 0.1077 & \textcolor[rgb]{0.753, 0.753, 0.753}{182.59} \\
            Autoformer \citep{wu2021autoformer} & 0.2562 & 0.2351 & 257.86 & 0.2896 & 0.3117 & 308.84 & 0.2235 & 0.1891 & 264.44 \\
            DLinear \citep{Zeng_Chen_Zhang_Xu_2023} & 0.1562 & 0.1061 & \textcolor[rgb]{0.804, 0.498, 0.196}{124.45} & 0.1557 & \textcolor[rgb]{0.804, 0.498, 0.196}{0.1063} & \textcolor[rgb]{0.753, 0.753, 0.753}{132.49} & 0.1538 & \textcolor[rgb]{0.804, 0.498, 0.196}{0.1050} & \textcolor[rgb]{0.804, 0.498, 0.196}{185.70} \\
            FEDformer \citep{pmlr-v162-zhou22g} & \textcolor[rgb]{0.753, 0.753, 0.753}{0.1357} & \textcolor[rgb]{0.753, 0.753, 0.753}{0.0706} & \textcolor[rgb]{0.804, 0.498, 0.196}{122.36} & \textcolor[rgb]{0.753, 0.753, 0.753}{0.1335} & \textcolor[rgb]{0.753, 0.753, 0.753}{0.0701} & \textcolor[rgb]{0.804, 0.498, 0.196}{143.40} & \textcolor[rgb]{0.753, 0.753, 0.753}{0.1341} & \textcolor[rgb]{0.753, 0.753, 0.753}{0.0700} & 194.76 \\
            FiLM \citep{NEURIPS2022_524ef58c} & \textcolor[rgb]{0.804, 0.498, 0.196}{0.1536} & 0.1108 & 136.99 & \textcolor[rgb]{0.804, 0.498, 0.196}{0.1528} & 0.1109 & 147.70 & \textcolor[rgb]{0.804, 0.498, 0.196}{0.1512} & 0.1096 & 211.73 \\
            \textbf{WECN (Ours)} & \textcolor[rgb]{1.0, 0.874, 0.0}{0.1257} & \textcolor[rgb]{1.0, 0.874, 0.0}{0.0701} & \textcolor[rgb]{1.0, 0.874, 0.0}{100.53} & \textcolor[rgb]{1.0, 0.874, 0.0}{0.1244} & \textcolor[rgb]{1.0, 0.874, 0.0}{0.0685} & \textcolor[rgb]{1.0, 0.874, 0.0}{121.10} & \textcolor[rgb]{1.0, 0.874, 0.0}{0.1174} & \textcolor[rgb]{1.0, 0.874, 0.0}{0.0679} & \textcolor[rgb]{1.0, 0.874, 0.0}{125.66} \\
            \bottomrule
        \end{tabular}}
\end{minipage}

\vspace{1.2em}

\begin{minipage}{\linewidth}
\centering
\textbf{(c) 60-step forecasting}

\vspace{0.5em}
\resizebox{\linewidth}{!}{

        \begin{tabular}{cccccccccc}
            \toprule
            \multicolumn{1}{c}{\multirow{2}{*}{BSL}} &
            \multicolumn{3}{c}{22m} &
            \multicolumn{3}{c}{23m} &
            \multicolumn{3}{c}{24m} \\
            \cmidrule(lr){2-4} \cmidrule(lr){5-7} \cmidrule(lr){8-10}
            \multicolumn{1}{c}{Metrics} & MAE & MSE & MAPE (\%) & MAE & MSE & MAPE (\%) & MAE & MSE & MAPE (\%) \\
            \midrule
            XGBoost \citep{chen2016xgboost} & 0.3584 & 0.2222 & 996.64 & 0.3601 & 0.2235 & 1024.54 & 0.3612 & 0.2244 & 966.57 \\
            LSTM \citep{hochreiter1997long} & 0.3327 & \textcolor[rgb]{0.753, 0.753, 0.753}{0.1683} & 1508.65 & 0.3582 & \textcolor[rgb]{0.753, 0.753, 0.753}{0.1705} & 1499.47 & 0.3503 & \textcolor[rgb]{0.753, 0.753, 0.753}{0.1651} & 1558.95 \\
            GRU \citep{chung2014empirical} & 0.3285 & 0.1792 & 1791.93 & 0.3238 & 0.1808 & 1508.45 & 0.3039 & \textcolor[rgb]{0.804, 0.498, 0.196}{0.1659} & 1620.99 \\
            Reformer \citep{Kitaev2020Reformer} & 0.2280 & 0.1831 & \textcolor[rgb]{0.753, 0.753, 0.753}{156.18} & 0.2273 & 0.1840 & \textcolor[rgb]{0.804, 0.498, 0.196}{181.79} & 0.2261 & 0.1868 & \textcolor[rgb]{0.804, 0.498, 0.196}{246.09} \\
            Autoformer \citep{wu2021autoformer} & 0.2714 & 0.2641 & 265.11 & 0.2912 & 0.2911 & 280.11 & 0.2622 & 0.2543 & 262.56 \\
            DLinear \citep{Zeng_Chen_Zhang_Xu_2023} & 0.2320 & 0.1927 & \textcolor[rgb]{0.804, 0.498, 0.196}{165.81} & 0.2319 & 0.1937 & \textcolor[rgb]{0.753, 0.753, 0.753}{180.28} & 0.2294 & 0.1920 & \textcolor[rgb]{1.0, 0.874, 0.0}{229.17} \\
            FEDformer \citep{pmlr-v162-zhou22g} & \textcolor[rgb]{1.0, 0.874, 0.0}{0.2113} & \textcolor[rgb]{0.804, 0.498, 0.196}{0.1760} & 174.93 & \textcolor[rgb]{0.753, 0.753, 0.753}{0.2093} & \textcolor[rgb]{0.804, 0.498, 0.196}{0.1767} & 195.22 & \textcolor[rgb]{0.753, 0.753, 0.753}{0.2072} & 0.1750 & 296.24 \\
            FiLM \citep{NEURIPS2022_524ef58c} & \textcolor[rgb]{0.804, 0.498, 0.196}{0.2209} & 0.1997 & 181.47 & \textcolor[rgb]{0.804, 0.498, 0.196}{0.2205} & 0.2010 & 199.22 & \textcolor[rgb]{0.804, 0.498, 0.196}{0.2186} & 0.1995 & 255.24 \\
            \textbf{WECN (Ours)} & \textcolor[rgb]{0.753, 0.753, 0.753}{0.2171} & \textcolor[rgb]{1.0, 0.874, 0.0}{0.1633} & \textcolor[rgb]{1.0, 0.874, 0.0}{134.49} & \textcolor[rgb]{1.0, 0.874, 0.0}{0.1967} & \textcolor[rgb]{1.0, 0.874, 0.0}{0.1534} & \textcolor[rgb]{1.0, 0.874, 0.0}{165.74} & \textcolor[rgb]{1.0, 0.874, 0.0}{0.1897} & \textcolor[rgb]{1.0, 0.874, 0.0}{0.1606} & \textcolor[rgb]{0.753, 0.753, 0.753}{229.90} \\
            \bottomrule
        \end{tabular}}
\end{minipage}

\captionsetup{justification=centering, font=small}
\footnotesize
    \begin{tablenotes}
        \item \footnotemark[1] BSL stands for Below Sea Level.
        \item \footnotemark[2] \textcolor[rgb]{1.0, 0.874, 0.0}{Gold} represents the best performance, \textcolor[rgb]{0.753, 0.753, 0.753}{Silver} represents the second best, \textcolor[rgb]{0.804, 0.498, 0.196}{Bronze} represents the third best.
    \end{tablenotes}

\label{tab:multistep}
\end{table*}

\begin{table}[htbp]
\centering
\caption{Tables of metrics for the ablation experiment of the DWT module.}
\label{tab:hyperparameter_search_space}
\resizebox{\columnwidth}{!}{
\begin{tabular}{@{}p{0.25\columnwidth}p{0.25\columnwidth}p{0.2\columnwidth}p{0.2\columnwidth}p{0.2\columnwidth}@{}}
\toprule
\textbf{Model} & \textbf{Data} & \textbf{MAE} & \textbf{MSE} & \textbf{MAPE (\%)}  \\
\midrule
\multirow{2}{*}{WECN-FFT}
& 23m & 0.0323 & 0.0043 & 30.95  \\
& $23\mathrm{m}_{\mathrm{new}}$ & 0.0371{\color{red} $\uparrow$} & 0.0115{\color{red} $\uparrow$} & 38.42{\color{red} $\uparrow$}  \\
\midrule
\multirow{2}{*}{WECN}
& 23m & 0.0245 & 0.0029 & 25.74  \\
& $23\mathrm{m}_{\mathrm{new}}$ & 0.0244{\color{green} $\downarrow$} & 0.0027{\color{green} $\downarrow$} & 25.38{\color{green} $\downarrow$}  \\
\bottomrule
\end{tabular}
}
\captionsetup{justification=centering, font=small}
\footnotesize
\begin{tablenotes}
    \item \footnotemark[1]{\color{red} $\uparrow$} represents an increase in forecasting error, and {\color{green} $\downarrow$} represents a decrease in forecasting error.
\end{tablenotes}
\label{88}
\end{table}

\begin{table}[htbp]
\centering
\caption{Tables of metrics for the ablation experiment of the TPE module.}
\resizebox{\columnwidth}{!}{
\begin{tabular}{@{}p{0.4\columnwidth}p{0.2\columnwidth}p{0.2\columnwidth}p{0.2\columnwidth}@{}}
\toprule
\textbf{Model} & \textbf{MAE} & \textbf{MSE} & \textbf{MAPE (\%)} \\
\midrule
FiLM w/o TPE & 0.0255 & 0.0033 & 29.15 \\
FiLM         & 0.0248 & \textbf{0.0029} & 27.66 \\
\midrule
FEDformer w/o TPE & 0.0429 & 0.0037 & 40.91 \\
FEDformer         & 0.0315 & 0.0032 & 31.20 \\
\midrule
WECN w/o TPE & 0.0810 & 0.0270 & 82.90 \\
WECN-RS     & 0.0650 & 0.0047 & 59.40 \\
WECN        & \textbf{0.0245} & \textbf{0.0029} & \textbf{24.74} \\
\bottomrule
\end{tabular}
}
\captionsetup{justification=centering, font=small}
\footnotesize
\begin{tablenotes}
    \item \footnotemark[1]The \textbf{bolded} metrics indicate the best performance across different models.
    \item \footnotemark[2] w/o stands for without.
\end{tablenotes}
\label{77}
\end{table}

In this research, we design experiments to analyze the interaction between multi-periodicity and forecasting accuracy in multi-step forecasting. The evaluation first examines the model's forecasting ability at 10-step, considering it a benchmark for short-term forecasting. Then, the analysis extends to longer multi-step forecasts, specifically the 30-step and 60-step. This progression allows us to examine whether and how the model's performance changes with increasing time steps.

\begin{figure}
\small
\centering
\includegraphics[width=8cm]{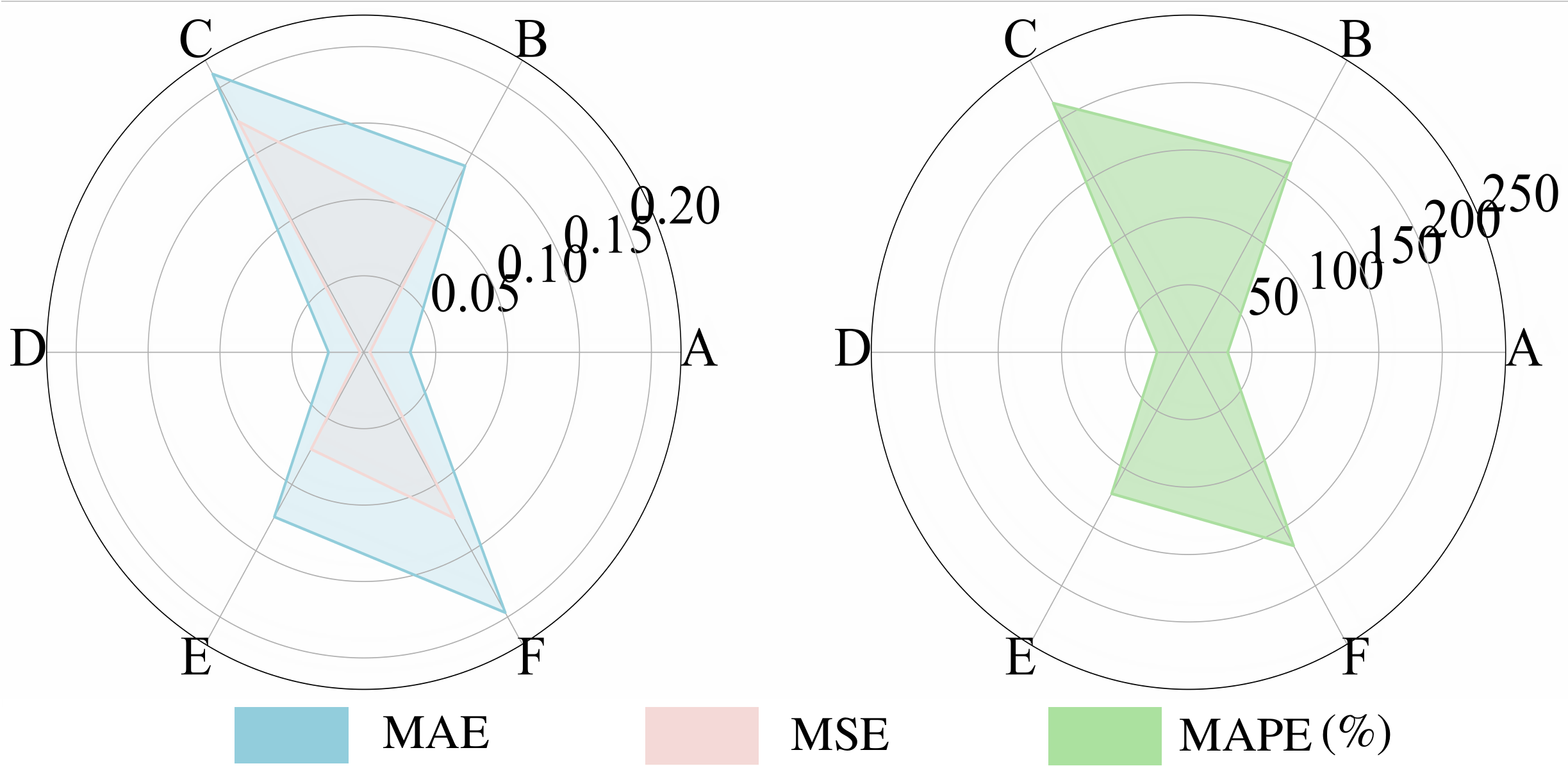}
\caption{Radar chart of the ablation experiment for the DWT module. A represents WECN-FFT with an output length of 10. B represents WECN-FFT with an output length of 30. C represents WECN-FFT with an output length of 60. D represents WECN with an output length of 10. E represents WECN with an output length of 30. F represents WECN with an output length of 60.} \label{8}
\end{figure}

Table~\ref{7}(a) presents the four evaluation metrics obtained under the 10-step forecasting setting using a stacked bar chart, while Table~\ref{7}(b) and Table~\ref{7}(c) summarize the results for the 30-step and 60-step forecasts, respectively.

Table~\ref{7}(b) and Table~\ref{7}(c) show that the reported values are averaged across different directions of tidal current speeds. Under the 10-step forecasting setting, WECN achieves an MAE of 0.0251, an MSE of 0.0028, and a MAPE of 36.24\%. Compared with the best-performing model other than WECN, the MAE and MSE decrease by 1.44\% and 14.29\%, respectively, while the MAPE increases by 2.11\%. Relative to the second-best model, the MAE and MSE are reduced by 27.06\% and 54.59\%, respectively, together with a 22.19\% reduction in MAPE. Overall, WECN achieves the best results in most cases, indicating that explicitly modeling the multi-periodic characteristics of tidal currents improves forecasting accuracy. In addition, the proposed framework maintains consistent performance across datasets collected at different water depths, demonstrating robust generalization ability.

Table~\ref{7}(b) and Table~\ref{7}(c) further indicate that forecasting performance degrades as the prediction horizon increases. In recursive multi-step forecasting, the prediction at the previous step is used as the input for the next step. Consequently, step-wise errors accumulate as the forecasting horizon grows, leading to reduced accuracy. Even when the horizon is extended to 30 and 60 steps, WECN alleviates performance degradation to a certain extent. This behavior is attributed to the architectural design of WECN, which converts the cascading temporal error accumulation common in one-dimensional models into a more controllable spatial error diffusion process, thereby improving stability in multi-step forecasting. As shown in Table~\ref{7}(b), for the 30-step forecasting task, compared with the best model other than WECN, MAE, MSE, and MAPE decrease by 8.88\%, 1.99\%, and 24.59\%, respectively. Relative to the second-best model, the reductions reach 21.09\% in MAE, 34.94\% in MSE, and 21.54\% in MAPE. Table~\ref{7}(c) shows that the proposed model continues to perform well in the 60-step forecasting task, particularly for the dataset collected at 23\,m below sea level. Among the three datasets, the 23\,m dataset exhibits the strongest non-stationarity. Since multi-periodicity includes non-stationary components within the time series, these results confirm that the proposed framework effectively learns multi-periodic characteristics.

\subsection{Ablation studies}
We conduct ablation experiments on the WECN framework to determine the impact of individual components on the framework's forecasting performance. Specifically, we examine the effects of the DWT and TPE modules.

\subsubsection{\texttimes{\textbf{Ablation study of the DWT module}}}

First, we validate the impact of the DWT module on forecasting performance. The DWT module was ablated. We used data at 23m below sea level to compare model performance across different output steps. The WECN without the DWT module uses the Fast Fourier Transform (FFT) to learn the periodic features of the sequence, which we refer to as WECN-FFT. Since the DWT module captures the local periodicity within the sequence, replacing it increases the forecasting error metrics for WECN-FFT. The results demonstrate the advantage of the DWT module in learning multi-periodicity. Figure \ref{8} shows the performance differences between the two periodicity-capturing methods.

To further verify the framework's ability to capture local periodicity, we added sinusoidal local periodicity to the original data. To control locality, a Gaussian window function was used. The Gaussian window was generated based on the relationship between time, the center offset, and the window width. Then, the Gaussian window was multiplied by a sine wave to create a locally periodic signal, which was added to the original time series data, resulting in a new dataset named $23\mathrm{m}_{\mathrm{new}}$. Next, the WECN and WECN-FFT models are used to forecast both the original and new datasets. Moreover, we optimize the hyperparameters. The performance metrics for forecasting are shown in Table \ref{88}.

The results indicate that the WECN-FFT model exhibits a decrease in prediction accuracy when dealing with data containing stronger local periodicity ($23\mathrm{m}_{\mathrm{new}}$). In contrast, the WECN framework improves prediction accuracy when handling the $23\mathrm{m}_{\mathrm{new}}$ data. The results validate the role of the DWT module. This further indicates that the high-frequency components retained in our data after being processed by the adaptive filter are meaningful and learnable patterns, not just aliased noise.

Results can be explained by the mismatch between global Fourier bases and non-stationary tidal dynamics. Real tidal-current sequences often contain quasi-periodic components with time-varying amplitude and phase, so the energy of a modulated sinusoid is not concentrated at a single discrete FFT bin but spreads to nearby frequencies due to finite-length truncation and windowing, which makes the extracted dominant periods sensitive to the chosen window and noise. By contrast, DWT implements a multiresolution filter bank that partitions the spectrum into dyadic subbands, and the periodicity strength can be measured by the subband coefficient energy. This representation is stable under mild frequency drift because the energy remains within the same subband even when the instantaneous frequency moves inside that band. Therefore, when locally windowed sinusoids are injected into $23\mathrm{m}_{\mathrm{new}}$, DWT yields a clearer separation between long-term tidal trends and short-term oscillatory bursts, whereas FFT-based features are more affected by spectral leakage and band spreading, which degrades WECN-FFT and improves WECN in the reported comparisons.

\subsubsection{\texttimes{\textbf{Ablation study of the TPE module}}}

After analyzing the contribution of the DWT module to the framework's predictive accuracy, we conducted the following comparative experiment to analyze further the impact of the hyperparameter optimization method on forecasting results:
\begin{itemize}
    \item \textbf{Using Random Sampler (i.e., WECN-RS):} We replaced the TPE parameter optimization method in the model with a random sampling method, which randomly selects hyperparameter values. The results showed a decline in predictive performance, indicating that the random sampling method could not effectively search for the optimal hyperparameter combination, thereby reducing the model's generalization performance.
    \item \textbf{Using Default Parameter Combinations (i.e., WECN w/o TPE):} We replaced the TPE parameter optimization method in the model with default parameter combinations. The results also showed a decline in forecasting performance, suggesting that the default parameter combinations could not adapt to different datasets and scenarios, limiting the model's flexibility.
\end{itemize}

To avoid restricting the analysis to the proposed model, the same TPE optimization procedure is also applied to two strong baseline models in this setting, namely FiLM and FEDformer, which show the second- and third-best performance among the compared baselines in Table~\ref{77}.

Table \ref{77} presents the forecasting performance of the model under different hyperparameter optimization methods. The results show that the model achieved the lowest metric values using the TPE parameter optimization method. The results confirm the effectiveness of the proposed method.

\section{Discussion}
\label{Section 4}
\subsection{Practical implications}
Accurate short-term tidal current speed forecasting serves as an operational module for tidal energy systems and grid integration. The proposed WECN models multi-periodicity arising from the interaction between tidal constituents and local flow conditions. This modelling design directly targets engineering deployment scenarios related to tidal energy integration into electrical grids. From a deployment perspective, the framework produces multi-step forecasts, which support operational decisions in cases where single-step prediction does not provide sufficient information for system controllability and reliability.

The study adopts a recursive multi-step forecasting strategy to generate a forecast trajectory. This trajectory supports:
\begin{itemize}
  \item short-horizon control actions, including set-point adjustment and power smoothing;
  \item longer-horizon operational planning, including scheduling and risk-aware decision-making based on an evolving future sequence rather than a single point estimate.
\end{itemize}
\subsection{Limitations}
Several limitations are associated with the current design. First, the current framework relies exclusively on tidal current time series and does not incorporate exogenous variables or knowledge-driven constraints. Accordingly, the integration of knowledge-driven approaches is identified as a direction for future work. In addition, the data processing pipeline applies missing-value imputation and predefined filtering operations, and hyperparameter selection is performed under a bounded evaluation budget. These design choices influence robustness and require further examination under varying data quality conditions and operational constraints.

 	\section{Conclusion}
\label{Section 5}
In this research, we developed a multi-periodicity learning framework that is capable of learning local periodicity. We extract dominant periods via DWT and feed them into the predictor without reconstructing components. Then, the TPE algorithm is integrated to optimize hyperparameters, forming the Wavelet-Enhanced Convolutional Network (WECN) framework for tidal current speed forecasting. Numerical results demonstrate that the proposed framework attains a 10-step average MAE of 0.025. Relative to the strongest baseline, error reductions range from 0.23\% to 6.66\%. In practical applications, tidal current time series exhibit multi-periodicity arising from the interaction of different tidal components and local flow conditions. The proposed framework models these multi-periodic features, enabling adaptation to the complex and dynamic patterns in real marine environments, which holds significant importance for practical engineering applications. Regarding generalization ability, the framework demonstrates applicability across different datasets and can be used for similar multi-periodicity time series tasks. The results indicate that the framework can provide valuable support for integrating tidal energy into the electrical grid. The results indicate that the framework can provide effective support for integrating tidal energy into the electrical grid. However, the current framework relies solely on tidal current time series, indicating the need to incorporate additional information and constraints in future work to extend the method.

	\section*{CRediT authorship contribution statement}
\textbf{Tengfei Cheng: }Methodology, Conceptualization, Software, Writing - original draft. \textbf{Yangdi Huang: }Data curation.  \textbf{Ling Xiao: }Writing - review \& editing. 
\textbf{Yunxuan Dong: }Conceptualization, Investigation, Writing - original draft, Writing - review \& editing.

\section*{Funding}

This work was supported in part by the Department of Science and Technology of Guangxi Zhuang Autonomous Region (grant number 2024JJB170087), in part by the Department of Human Resources and Social Security of Guangxi Zhuang Autonomous Region (grant number 202401950).

	\section*{Declaration of competing interest}
The authors declare that they have no known competing financial interests or personal relationships that could have appeared to influence the work reported in this paper.

	\section*{Data availability}
    Data will be made available on request.

	\bibliographystyle{model5-names}

	\bibliography{cas-refs}

@article{CAGLAYAN2019113794,
title = {The techno-economic potential of offshore wind energy with optimized future turbine designs in Europe},
journal = {Appl. Energy},
volume = {255},
pages = {113794},
year = {2019},
issn = {0306-2619},
author = {Dilara Gulcin Caglayan and David Severin Ryberg and Heidi Heinrichs and Jochen Linßen and Detlef Stolten and Martin Robinius}
}

@article{qian2022tidal,
  title={Tidal current prediction based on a hybrid machine learning method},
  author={Qian, Peng and Feng, Bo and Liu, Xiaodong and Zhang, Dahai and Yang, Jing and Ying, You and Liu, Cong and Si, Yulin},
  journal={Ocean Eng.},
  volume={260},
  pages={111985},
  year={2022},
  publisher={Elsevier}
}

@article{doodson1957analysis,
  title={The analysis and prediction of tides in shallow water},
  author={Doodson, Arthur Thomas},
  journal={IHR},
  year={1957}
}

@INPROCEEDINGS{5589321,
  author={Jahromi, Mahda and Maswood, Ali I. and Tseng, K.J.},
  booktitle={IEEE PES General Meeting}, 
  title={Comparison of different techniques for short term prediction of tidal current speeds}, 
  year={2010},
  volume={},
  number={},
  pages={1-8},
  doi={10.1109/PES.2010.5589321}}

@article{KIM2024118010,
title = {Techno-economic analysis for design and management of international green hydrogen supply chain under uncertainty: An integrated temporal planning approach},
journal = {Energy Convers. Manage.},
volume = {301},
pages = {118010},
year = {2024},
issn = {0196-8904},
author = {Sunwoo Kim and Joungho Park and Wonsuk Chung and Derrick Adams and Jay H. Lee}
}

@article{ZHANG2022611,
title = {Short-term offshore wind power forecasting - A hybrid model based on {Discrete Wavelet Transform (DWT)}, {Seasonal Autoregressive Integrated Moving Average (SARIMA)}, and deep-learning-based {Long Short-Term Memory (LSTM)}},
journal = {Renew. Energy},
volume = {185},
pages = {611-628},
year = {2022},
issn = {0960-1481},
author = {Wanqing Zhang and Zi Lin and Xiaolei Liu}
}

@inproceedings{wu2022timesnet,
  title={Timesnet: Temporal 2d-variation modeling for general time series analysis},
  author={Wu, Haixu and Hu, Tengge and Liu, Yong and Zhou, Hang and Wang, Jianmin and Long, Mingsheng},
  booktitle={ICLR},
  year={2022}
}

@article{YANG2024119603,
title = {A review of tidal current power generation farm planning: Methodologies, characteristics and challenges},
journal = {Renew. Energy},
volume = {220},
pages = {119603},
year = {2024},
issn = {0960-1481},
author = {Zhixue Yang and Zhouyang Ren and Hui Li and Zhen Pan and Weiyi Xia}
}

@book{Darwin_2009,
  place={Cambridge},
  series={Cambridge Library Collection - Physical Sciences},
  title={The Scientific Papers of Sir George Darwin: Tidal Friction and Cosmogony},
  publisher={Cambridge University Press},
  author={Darwin, George Howard},
  year={2009},
  collection={Cambridge Library Collection - Physical Sciences}
}

@article{Fornerino_LeProvost_2015,
  title={A Model for Prediction of the Tidal Currents in the English Channel},
  volume={62},
  abstractNote={A model for prediction of tidal currents in the English Channel is presented. It is based on the classical harmonic description of tides deduced from the spectral development of the luni-solar tidal potential. The spatial distribution of the characteristic parameters (intensity, phase, and direction of the maximum velocity vector, ellipticity of the hodograph) for the 26 harmonic constituents introduced in the prediction procedure are deduced from a numerical simulation of 1 month’s duration for the entire English Channel. Two kinds of documents can be produced from this model : instantaneous velocity fields over a given area, and time series of the intensity and the direction of the velocity vector at a given location, over a given period. Four examples of prediction are presented, corresponding to specific areas and over periods where tidal currents have actually been observed. The comparison between predictions and observations is very satisfactory.},
  number={2},
  journal={IHR},
  author={Fornerino, M. and Le Provost, C.},
  year={2015},
  month={Jul.}
}

@ARTICLE{9169644,
  author={Yang, Cheng-Hong and Wu, Chih-Hsien and Hsieh, Chih-Min},
  journal={IEEE Access}, 
  title={Long Short-Term Memory Recurrent Neural Network for Tidal Level Forecasting}, 
  year={2020},
  volume={8},
  number={},
  pages={159389-159401},
 }

@article{GRABBE20091898,
title = {A review of the tidal current energy resource in Norway},
journal = {Renew. Sustain. Energy Rev.},
volume = {13},
number = {8},
pages = {1898-1909},
year = {2009},
issn = {1364-0321},
doi = {https://doi.org/10.1016/j.rser.2009.01.026},
author = {Mårten Grabbe and Emilia Lalander and Staffan Lundin and Mats Leijon}
}

@article{MONAHAN2023103596,
title = {A hybrid model for online short-term tidal energy forecasting},
journal = {Appl. Ocean Res.},
volume = {137},
pages = {103596},
year = {2023},
issn = {0141-1187},
author = {Thomas Monahan and Tianning Tang and Thomas A.A. Adcock}
}

@article{SHAO2023125476,
title = {A decision framework for tidal current power plant site selection based on {GIS-MCDM}: A case study in China},
journal = {Energy},
volume = {262},
pages = {125476},
year = {2023},
issn = {0360-5442},
author = {Meng Shao and Yuanxu Zhao and Jinwei Sun and Zhixin Han and Zhuxiao Shao}
}

@article{owen1980three,
  title={A three-dimensional model of the Bristol Channel},
  author={Owen, A},
  journal={J. Phys. Oceanogr.},
  volume={10},
  number={8},
  pages={1290--1302},
  year={1980},
  publisher={American Meteorological Society}
}

@inproceedings{sarkar2016machine,
  title={A Machine Learning Approach to the Prediction of Tidal Currents,},
  author={Sarkar, Dripta and Osborne, Michael and Adcock, Thomas},
  booktitle={ISOPE},
  pages={ISOPE-I-16-514},
  year={2016},
  organization={ISOPE}
}

@article{sarkar2018prediction,
  title={Prediction of tidal currents using Bayesian machine learning},
  author={Sarkar, Dripta and Osborne, Michael A and Adcock, Thomas AA},
  journal={Ocean Eng.},
  volume={158},
  pages={221--231},
  year={2018},
  publisher={Elsevier}
}

@article{jay1999comparison,
  title={A comparison of methods for analysis of tidal records containing multi-scale non-tidal background energy},
  author={Jay, David A and Flinchem, Edward P},
  journal={Cont. Shelf Res.},
  volume={19},
  number={13},
  pages={1695--1732},
  year={1999},
  publisher={Elsevier}
}

@inproceedings{wang2023wavelet,
  title={WHEN: A Wavelet-DTW hybrid attention network for heterogeneous time series analysis},
  author={Wang, Jingyuan and Yang, Chen and Jiang, Xiaohan and Wu, Junjie},
  booktitle={Proceedings of the 29th ACM SIGKDD Conference on Knowledge Discovery and Data Mining},
  pages={2361--2373},
  year={2023}
}

@inproceedings{zhou2022fedformer,
  title={Fedformer: Frequency enhanced decomposed transformer for long-term series forecasting},
  author={Zhou, Tian and Ma, Ziqing and Wen, Qingsong and Wang, Xue and Sun, Liang and Jin, Rong},
  booktitle={ICML},
  pages={27268--27286},
  year={2022},
  organization={PMLR}
}

@inproceedings{he2016deep,
  title={Deep residual learning for image recognition},
  author={He, Kaiming and Zhang, Xiangyu and Ren, Shaoqing and Sun, Jian},
  booktitle={CVPR},
  pages={770--778},
  year={2016}
}

@INPROCEEDINGS{7298594,
  author={Szegedy, Christian and Wei Liu and Yangqing Jia and Sermanet, Pierre and Reed, Scott and Anguelov, Dragomir and Erhan, Dumitru and Vanhoucke, Vincent and Rabinovich, Andrew},
  booktitle={CVPR}, 
  title={Going deeper with convolutions}, 
  year={2015},
  volume={},
  number={},
  pages={1-9},
  doi={10.1109/CVPR.2015.7298594}}

@inproceedings{NEURIPS2021_bcc0d400,
 author = {Wu, Haixu and Xu, Jiehui and Wang, Jianmin and Long, Mingsheng},
 booktitle = {Adv. Neural Inf. Process. Syst.},
 editor = {M. Ranzato and A. Beygelzimer and Y. Dauphin and P.S. Liang and J. Wortman Vaughan},
 pages = {22419--22430},
 publisher = {Curran Associates, Inc.},
 title = {Autoformer: Decomposition Transformers with Auto-Correlation for Long-Term Series Forecasting},
 volume = {34},
 year = {2021}
}

@misc{tiger2022cost,
  author       = {{Tidal Stream Industry Energiser (TIGER) Project}},
  title        = {Cost Reduction Pathway of Tidal Stream Energy in the UK and France},
  year         = {2022},
  month        = {October},
  version      = {1.0},
  url          = {https://ore.catapult.org.uk/resource-hub/analysis-reports/cost-reduction-pathway-of-tidal-stream-energy-in-the-uk-and-france},
  note         = {Accessed: 2024-07-30},
  institution  = {European Regional Development Fund}
}

@article{ALISALAMAI2023118658,
title = {Deep learning framework for predictive modeling of crude oil price for sustainable management in oil markets},
journal = {Expert Syst. Appl.},
volume = {211},
pages = {118658},
year = {2023},
issn = {0957-4174},
author = {Abdullah {Ali Salamai}}
}

@article{Chavat2022ECDUY,
  title        = {ECD-UY, detailed household electricity consumption dataset of Uruguay},
  author       = {Chavat, Joaquín and Nesmachnow, Sergio and Graneri, Juan and others},
  journal      = {Scientific Data},
  volume       = {9},
  pages        = {21},
  year         = {2022},
  publisher    = {Nature Publishing Group},
  doi          = {10.1038/s41597-022-01122-x}
}

@inproceedings{Kitaev2020Reformer,
  title = {Reformer: The Efficient Transformer},
  author = {Nikita Kitaev and Lukasz Kaiser and Anselm Levskaya},
  booktitle = {International Conference on Learning Representations},
  year = {2020}
}

@inproceedings{wu2021autoformer,
  title = {Autoformer: Decomposition Transformers with Auto-Correlation for Long-Term Series Forecasting},
  author = {Haixu Wu and Jiehui Xu and Jianmin Wang and Mingsheng Long},
  booktitle = {Advances in Neural Information Processing Systems},
  editor = {A. Beygelzimer and Y. Dauphin and P. Liang and J. Wortman Vaughan},
  year = {2021}
}

@article{Zeng_Chen_Zhang_Xu_2023,
  title = {Are Transformers Effective for Time Series Forecasting?},
  volume = {37},
  DOI = {10.1609/aaai.v37i9.26317},
  number = {9},
  journal = {Proceedings of the AAAI Conference on Artificial Intelligence},
  author = {Zeng, Ailing and Chen, Muxi and Zhang, Lei and Xu, Qiang},
  year = {2023},
  month = {Jun.},
  pages = {11121-11128}
}

@InProceedings{pmlr-v162-zhou22g,
  title = {{FED}former: Frequency Enhanced Decomposed Transformer for Long-term Series Forecasting},
  author = {Zhou, Tian and Ma, Ziqing and Wen, Qingsong and Wang, Xue and Sun, Liang and Jin, Rong},
  booktitle = {Proceedings of the 39th International Conference on Machine Learning},
  pages = {27268--27286},
  year = {2022},
  editor = {Chaudhuri, Kamalika and Jegelka, Stefanie and Song, Le and Szepesvari, Csaba and Niu, Gang and Sabato, Sivan},
  volume = {162},
  series = {Proceedings of Machine Learning Research},
  month = {17--23 Jul},
  publisher = {PMLR}
}

@inproceedings{NEURIPS2022_524ef58c,
  author = {Zhou, Tian and Ma, Ziqing and Wang, Xue and Wen, Qingsong and Sun, Liang and Yao, Tao and Yin, Wotao and Jin, Rong},
  booktitle = {Advances in Neural Information Processing Systems},
  editor = {S. Koyejo and S. Mohamed and A. Agarwal and D. Belgrave and K. Cho and A. Oh},
  pages = {12677--12690},
  publisher = {Curran Associates, Inc.},
  title = {FiLM: Frequency Improved Legendre Memory Model for Long-term Time Series Forecasting},
  volume = {35},
  year = {2022}
}

@article{jia2024witran,
  title={WITRAN: Water-wave information transmission and recurrent acceleration network for long-range time series forecasting},
  author={Jia, Yuxin and Lin, Youfang and Hao, Xinyan and Lin, Yan and Guo, Shengnan and Wan, Huaiyu},
  journal={Adv. Neural Inf. Process. Syst.},
  volume={36},
  year={2024}
}

@article{wang2024tssurvey,
  title={Deep Time Series Models: A Comprehensive Survey and Benchmark},
  author={Yuxuan Wang and Haixu Wu and Jiaxiang Dong and Yong Liu and Mingsheng Long and Jianmin Wang},
  journal={arXiv preprint arXiv:2407.13278},
  year={2024},
}

@article{hochreiter1997long,
  title={Long short-term memory},
  author={Hochreiter, Sepp and Schmidhuber, J{\"u}rgen},
  journal={Neural computation},
  volume={9},
  number={8},
  pages={1735--1780},
  year={1997},
  publisher={MIT Press}
}

@article{chung2014empirical,
  title={Empirical evaluation of gated recurrent neural networks on sequence modeling},
  author={Chung, Junyoung and Gulcehre, Caglar and Cho, KyungHyun and Bengio, Yoshua},
  journal={arXiv preprint arXiv:1412.3555},
  year={2014}
}

@inproceedings{chen2016xgboost,
  title={XGBoost: A Scalable Tree Boosting System},
  author={Chen, Tianqi and Guestrin, Carlos},
  booktitle={Proceedings of the 22nd ACM SIGKDD International Conference on Knowledge Discovery and Data Mining},
  pages={785--794},
  year={2016}
}

@article{OROURKE2014726,
title = {Ireland’s tidal energy resource; An assessment of a site in the Bulls Mouth and the Shannon Estuary using measured data},
journal = {Energy Convers. Manage.},
volume = {87},
pages = {726-734},
year = {2014},
issn = {0196-8904},
author = {Fergal O’Rourke and Fergal Boyle and Anthony Reynolds}
}

@article{du2019novel,
  title={A novel hybrid model for short-term wind power forecasting},
  author={Du, Pei and Wang, Jianzhou and Yang, Wendong and Niu, Tong},
  journal={Appl. Soft Comput.},
  volume={80},
  pages={93--106},
  year={2019},
  publisher={Elsevier}
}

@article{YANG2022119849,
title = {A fuzzy intelligent forecasting system based on combined fuzzification strategy and improved optimization algorithm for renewable energy power generation},
journal = {Appl. Energy},
volume = {325},
pages = {119849},
year = {2022},
issn = {0306-2619},
author = {Hufang Yang and Ping Jiang and Ying Wang and Hongmin Li}
}

@article{MA2020120159,
title = {Research and application of association rule algorithm and an optimized grey model in carbon emissions forecasting},
journal = {Technol. Forecast. Soc. Change},
volume = {158},
pages = {120159},
year = {2020},
issn = {0040-1625},
author = {Xuejiao Ma and Ping Jiang and Qichuan Jiang}
}

@ARTICLE{6665108,
  author={Wan, Can and Xu, Zhao and Pinson, Pierre and Dong, Zhao Yang and Wong, Kit Po},
  journal={IEEE Trans. Power Syst.}, 
  title={Probabilistic Forecasting of Wind Power Generation Using Extreme Learning Machine}, 
  year={2014},
  volume={29},
  number={3},
  pages={1033-1044}}

@article{Li2024,
  author    = {Y. Li and Y. Ding and S. He and others},
  title     = {Artificial intelligence-based methods for renewable power system operation},
  journal   = {Nat. Rev. Electr. Eng.},
  volume    = {1},
  pages     = {163--179},
  year      = {2024},
  doi       = {10.1038/s44229-024-00029-4},
}




\end{sloppypar}
\end{document}